\documentclass{IEEEtran}
\usepackage{graphicx}
\usepackage{cite}
\usepackage{picinpar}
\usepackage{amsmath}
\usepackage{amssymb}
\usepackage{url}
\usepackage{flushend}
\usepackage[latin1]{inputenc}
\usepackage{colortbl}
\usepackage{soul}
\usepackage{multirow}
\usepackage{pifont}
\usepackage{color}
\usepackage{alltt}
\usepackage[hidelinks]{hyperref}
\usepackage{enumerate}
\usepackage{siunitx}
\usepackage{breakurl}
\usepackage{epstopdf}
\usepackage{pbox}
\usepackage{subcaption}
\usepackage[most]{tcolorbox}
\usepackage{float}

\begin{document}
\title{3D Dense Face Alignment with Fused Features by Aggregating CNNs and GCNs}

\author{
	\vskip 1em
	
	Yanda~Meng, Xu~Chen, Dongxu~Gao, Yitian~Zhao, Xiaoyun~Yang, Yihong~Qiao, Xiaowei~Huang and Yalin~Zheng*
     
	\thanks{
	
		%Manuscript received Month xx, 2xxx; revised Month xx, xxxx; accepted Month x, xxxx.
		This work was supported by the China Science IntelliCloud Technology Co., Ltd.
		
		Y. Meng, X. Chen, D. Gao, and Y. Zheng are with the Institute of Life Course and Medical Sciences, University of Liverpool, Liverpool, L7 8TX, United Kingdom.
		
		Y. Zhao is with the Cixi Institute of Biomedical Engineering, Ningbo Institute of Materials Technology and Engineering, Chinese Academy of Sciences, Ningbo 315201, China.
		
		X. Huang is with the Department of Computer Science, University of Liverpool, Liverpool, L7 8TX, United Kingdom.
		
		Y. Qiao is with the China Science IntelliCloud Technology Co., Ltd, Shanghai, China.
		
		X. Yang is with the Remark AI UK Limited, London, SE1 9PD, United Kingdom.
		
		Corresponding author: yalin.zheng@liverpool.ac.uk.
	}
}

\maketitle
	
\begin{abstract}
In this paper, we propose a novel multi-level aggregation network to regress the coordinates of the vertices of a 3D face from a single 2D image in an end-to-end manner. This is achieved by seamlessly combining standard convolutional neural networks (CNNs) with Graph Convolution Networks (GCNs). By iteratively and hierarchically fusing the features across different layers and stages of the CNNs and GCNs, our approach can provide a dense face alignment and 3D face reconstruction simultaneously for the benefit of direct feature learning of 3D face mesh. Experiments on several challenging datasets demonstrate that our method outperforms state-of-the-art approaches on both 2D and 3D face alignment tasks.
\end{abstract}


\markboth{}%
{}

\section{Introduction}

Face alignment and 3D face reconstruction are two interrelated problems in the field of computer vision and graphics research and industrial applications. Face alignment aims to locate specific 2D face landmarks, which is essential for most facial image applications such as face recognition  \cite{wagner2011toward}, facial expression recognition  \cite{jeni2014spatio} or head pose analysis  \cite{derkach2017head}. However, problems such as occlusions, large pose, and extreme lighting conditions make it a difficult task. In the past decades, researchers started to solve face alignment problems through 3D facial reconstruction by exploring the strong correlations between 2D landmarks and 3D faces. Since the introduction of 3D Morphable Model (3DMM) in 1999  \cite{blanz1999morphable}, several methods have been proposed to extend it to restore a 3D face mesh from a 2D facial image \cite{tran2018nonlinear,tuan2017regressing,jackson2017large,dou2017end}, which can provide both 3D face reconstruction and dense face alignment results.  More recently, Convolution Neural Networks (CNNs) have been developed and used to directly regress the parameters of 3DMM model from images \cite{zhu2016face,jourabloo2016large,richardson2017learning}. However, the performance of these model-based methods are still limited by the face reconstruction from a low-dimensional subspace of parametric 3DMM model.

\begin{figure}[t] %pic result
	\centering
	\begin{minipage}[t]{0.48\textwidth}
		\centering
		\includegraphics[width=8cm]{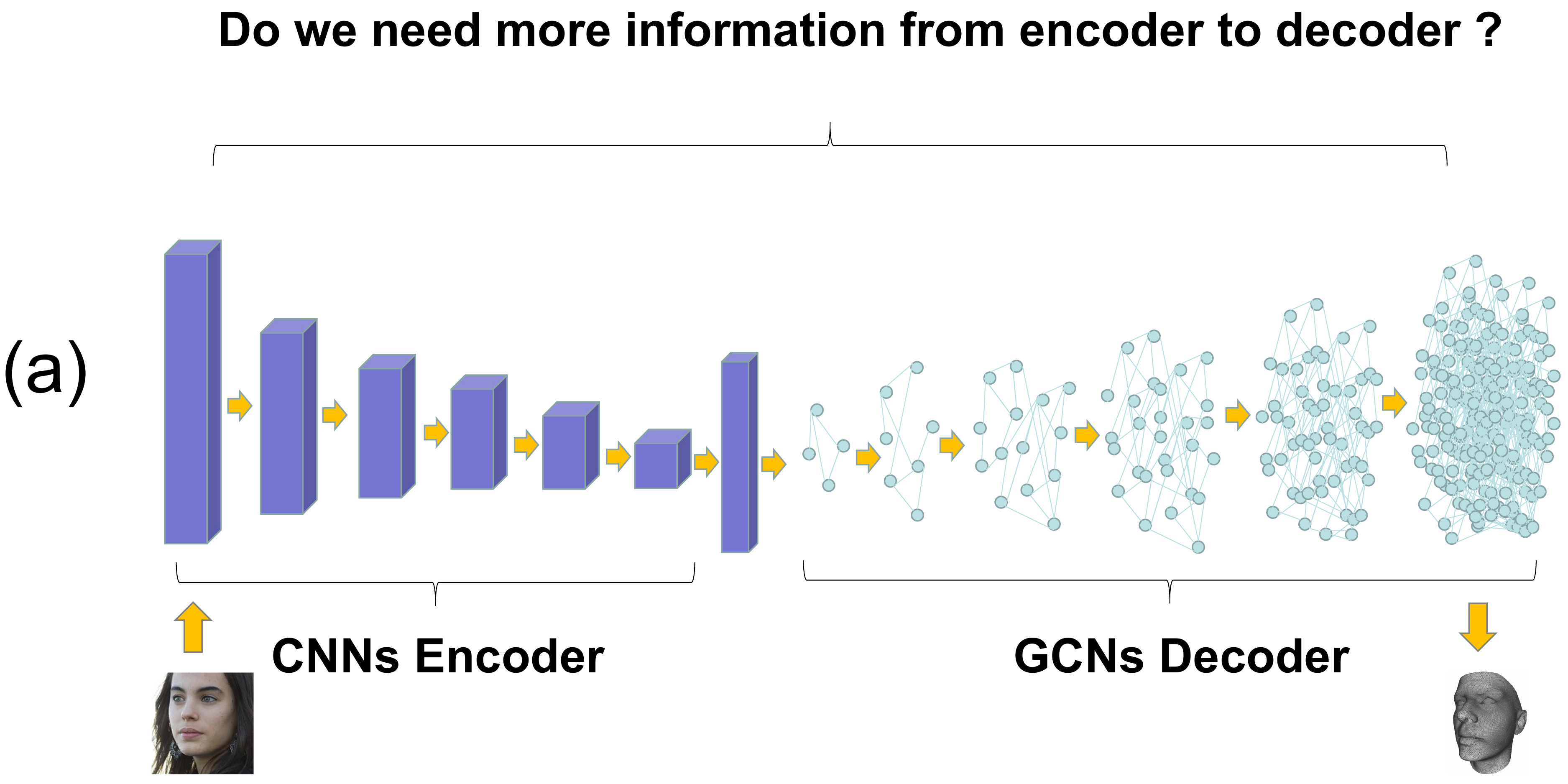}
	\end{minipage}
	%\vspace{2em}
	\begin{minipage}[t]{0.48\textwidth}
		\centering
		\includegraphics[width=8cm]{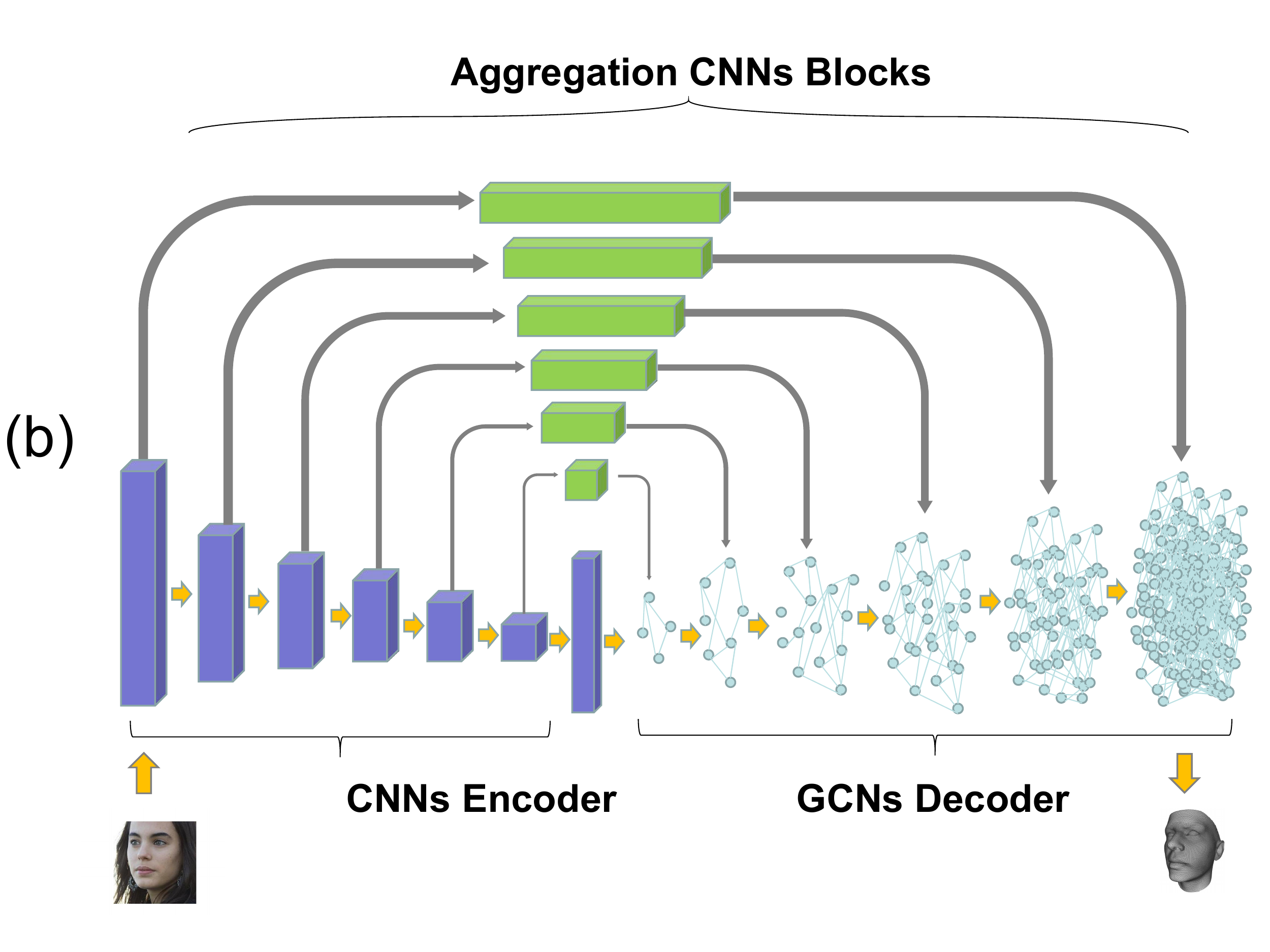}
		
	\end{minipage}
	\caption{Diagrams illustrating the difference between a mesh encoder-decoder and our proposed method. (a) An encoder-decoder structure used by existing methods  \cite{zhou2019dense} to regress 3D face mesh from latent embeddings. (b) Our method. As illustrated, our model fuses and reuses multi-level spatial and semantic features from an input face, which works as extra input information to help GCNs decoder to reconstruct the coordinates of face vertices better.}
	\label{1}
\end{figure}

To address this problem, different strategies have been proposed by using the most recent deep learning methods to regress the 3D face coordinates from 2D representations, such as Projected  Coordinate Code (PNCC)  \cite{richardson2017learning}, quantized conformal mapping  \cite{alp2017densereg}, depth images  \cite{sela2017unrestricted} and conformal UV maps \cite{feng2018joint}. Although these methods can regress the 3D geometry from 2D representations, their performance is often susceptible to the noise introduced by the 2D representation process from isolated mesh points. 

Graph Convolution Networks (GCNs) have recently shown great potential to tackle non-grid like data such as 3D face meshes \cite{monti2017geometric}. If it is used to perform convolution on 3D meshes directly, it will necessitate 2D representations as required by the previous methods and thus reduce (or avoid) noise in the 2D representation. CoMA  \cite{ranjan2018generating} proposes a mesh encoder-decoder to learn a non-linear representation on the 3D face surface and reconstructs the 3D face mesh via GCNs. Following CoMA,  \cite{zhou2019dense} propose an encoder-decoder network, which encodes input images into latent embeddings then decodes the embeddings to 3D face mesh with GCNs. We believe that, during the encoders downsampling process, some content information from face image will be lost. As for the decoder, the only input is the latent embeddings, which cannot adequately represent low-level semantic information and high-level spatial image features of the input face.

In this paper, we propose an end-to-end approach that directly learns multi-level regression mappings from image pixels to 3D face mesh vertices by seamlessly combining CNNs and GCNs for 3D face alignment and reconstruction. In this model, we perform feature learning on face meshes and utilize additional multi-level features fused from the input image in a hierarchical manner that helps GCNs regress more accurate 3D face vertices. Our model attains superior performance on 2D and 3D face alignment tasks to state-of-the-art methods. In particular, our model outperforms other methods by a large margin on the large pose face alignment problem, because with the help of aggregative feature learning, our model gains more useful information from visible parts of the input face image, which helps GCNs better regressing the invisible mesh vertices. Our model is light-weight and only needs 16.0 ms to provide 3D face vertices on a test image.

\subsection{Contributions}
Our approach works well with all kinds of face images, including arbitrary poses, facial expressions and occlusions. The contributions of our paper are as follows:

1.) To the best of our knowledge, this is the first time that 3D facial geometry is directly recovered from 2D images in an end-to-end fashion through fusing features from different levels enabled by connections between CNNs and GCNs. We demonstrate that low-level semantic information and the high-level spatial feature can be fully utilized to estimate 3D facial geometry. This is different from the recently proposed encoder-decoder networks \cite{zhou2019dense}, which only use low-level latent embeddings.
2.) We propose a novel light-weight and efficient aggregation network to regress more accurate 3D face mesh vertices from corresponding in-the-wild 2D facial images. For training, we propose a new loss function for facial landmarks localization, which helps to prevent taking large update steps when approaching a small range of errors in the late training stage. 
3.) Comprehensive experiments have been undertaken on several challenging datasets to evaluate the performance of the new model. The quantitative and qualitative results confirmed its superiority to other state-of-the-art approaches. In particular, our model outperforms previous methods on 2D and 3D large pose face alignment tasks by more than 18\% relative improvement. %alleged aggregation \cite{zhou2018unet++,yu2018deep} structure

\section{Related work}%\vspace{-0.1em}
%To better understand the problem of dense face alignment and 3D face reconstruction, we provide an in-depth review of related methods.

\subsection{3D Morphable Models}%\vspace{-0.1em}% 3dmm
3DMM is an affine parametric model of face geometry where the texture is learned from high-quality face scans \cite{blanz1999morphable}. It is a PCA-based implementation that produces new shape instances from a combination of linear bases of the training images. Recent approaches  \cite{zhu2016face,jourabloo2016large,dou2017end,jourabloo2015pose,richardson20163d} can be seen as an extension of 3DMM by estimating the 3DMM parameters by using CNN networks in a supervised manner. \cite{zhu2016face,jourabloo2016large,richardson20163d} proposed using cascaded CNNs to approximate the non-linear optimization function and to regress the 3DMM parameters iteratively. They demonstrated the effectiveness of CNNs in solving the complex mapping function from a 2D face image to 3DMM parameters, but it took a long time to train the network due to the iterations. %Both \cite{zhu2016face,richardson20163d} use color images and 3D renderings of an initial 3D face mesh as input to the network, while Richardson \textit{et al.} \cite{richardson20163d} also use the initial random 3DMM parameter as input to build a feedback mechanism and supervise the CNNs to update the 3DMM parameters iteratively.
\cite{dou2017end,jourabloo2015pose,tuan2017regressing} proposed end-to-end CNNs to directly estimate the 3DMM parameters. In particular, \cite{tuan2017regressing} used a very deep CNN to regress shape and texture parameters of 3DMM for 3D face recognition to improve the discriminative identity of reconstructed face meshes. Other methods like \cite{booth2018large,booth20173d,gecer2019ganfit,gecer2018semi,deng2018uv} focused on optimization-based texture generation methods, for example, Booth \textit{et al.} \cite{booth2018large} used 3DMM fits to in-the-wild images and Principal Component Pursuit with missing values to complete the unobserved texture. Both \cite{gecer2019ganfit,deng2018uv} employed Generative Adversarial Networks (GANs) to learn a powerful generator of facial texture, in particular, Gecer \textit{et al.} \cite{gecer2019ganfit} used differentiable rendering layer to self-supervise model to learn the texture information.
%Although these above model-based methods achieve promising results on face alignment tasks, they are limited by the capacity of the underlying 3DMM model, and also a lack of 3D spatial information and geometry correspondence during training.

\subsection{Geometric Deep Learning}%\vspace{-0.1em}% Convolutional Networks on Graphs
GCNs have shown their superior ability on several computer vision tasks such as scene understanding \cite{Meng_2022_Adaptive_Counting,meng2021spatial}, image segmentation \cite{meng2020regression,meng2021bi,meng2020cnn,meng2021graph}, \textit{etc..}
CNNs are effective on Euclidean data such as images but not good at non-Euclidean domains such as grids in face mesh \cite{bronstein2017geometric}. To overcome the disadvantages of CNNs, GCNs from geometric deep learning have recently been proposed. Bruna \textit{et al.} \cite{bruna2013spectral} proposed convolutions in the spectral domain defined by the eigenvectors of Laplacian graphs whereas the filters were parametrized with a smooth transfer function. Still, it is expensive to compute and unable to extract low-level features on the graph.
%However, Levie \textit{et al.} \cite{levie2018cayleynets} solved this problem with parametric Cayley polynomials functions to capture narrow frequency bands without Laplacian eigendecomposition, which reduces computational complexity and obtains better spectral resolution results. 
ChebyNet \cite{defferrard2016convolutional} solved the computational complexity problem with Chebyshev polynomial functions, which directly applied it to Laplacian graphs without computing the Fourier basis. CoMA \cite{ranjan2018generating} applied ChebyNet to 3D face meshes to find a low-dimensional non-linear representation of faces with an encoder-decoder structure. By spectral graph convolution and mesh sampling operations, it achieves state-of-the-art results in 3D face mesh generation.

\subsection{Aggregation Network}%\vspace{-0.1em}
Aggregation networks have shown powerful ability in visual recognition tasks because these tasks require rich information that spans channels or depth, scales, and resolutions \cite{yu2018deep}.

Densely connected networks (DenseNets) \cite{huang2017densely} aggregated across channels and depths. It improved the induction of recognition through propagating features and losses from skip connections, which concatenated every layer in stages. Feature pyramid networks (FPNs)  \cite{lin2017feature} aggregated features across different resolutions and scales. It restricted features through adjusting resolutions and semantics and aggregated over the degrees of a pyramidal component progressive system by top-down and parallel associations. 
%Other methods like  \cite{yu2018deep,zhou2018unet++}, propose aggregation networks that can fuse features in both semantic and spatial levels. DLA \cite{yu2018deep} employs a deep layer aggregation network which outperforms in several visual tasks, including classification, segmentation, and recognition. UNet++ \cite{zhou2018unet++} uses a similar idea as DLA and surpasses U-Net  \cite{ronneberger2015u} in medical image segmentation tasks. Still, it also retains the skip-connection structure to reduce the semantic gap between sub-networks. 
Instead of a skip-connection design, RefineNet  \cite{lin2017refinenet} introduced a refine module to extract the multi-scale features between encoder and decoder. MCUA  \cite{nie2019multi} used multi-level context ultra-aggregation to combine intra and inter level features for stereo matching. Likewise, DFANet  \cite{li2019dfanet} aggregated discriminative features through sub-networks and sub-stages cascade, respectively. RefineNet, MCUA, and DFANet showed good performance on 2D semantic segmentation through aggregating features. Compared to these methods, our proposed aggregation block can fuse and reuse multi-level features iteratively and hierarchically across different layers and stages. Our model combines CNNs and GCNs, solving 2D to 3D face reconstruction and dense face alignment task simultaneously.
\begin{figure*}[t] %pic result
    \centering
    \begin{minipage}[t]{0.13\textwidth}
        \centering
        \includegraphics[width=2.7cm]{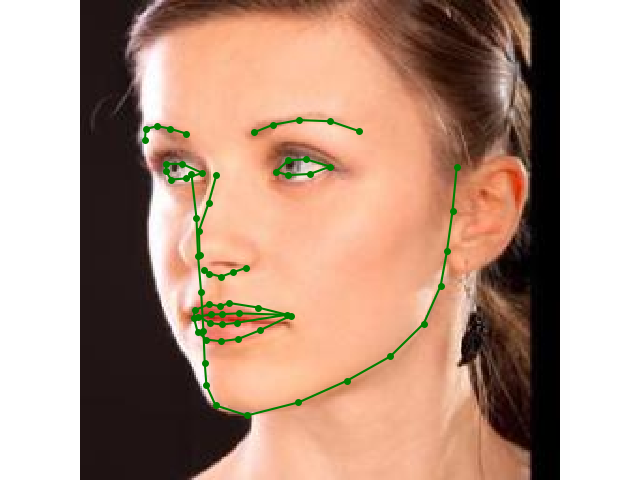}
    \end{minipage}
    \begin{minipage}[t]{0.13\textwidth}
        \centering
        \includegraphics[width=2.7cm]{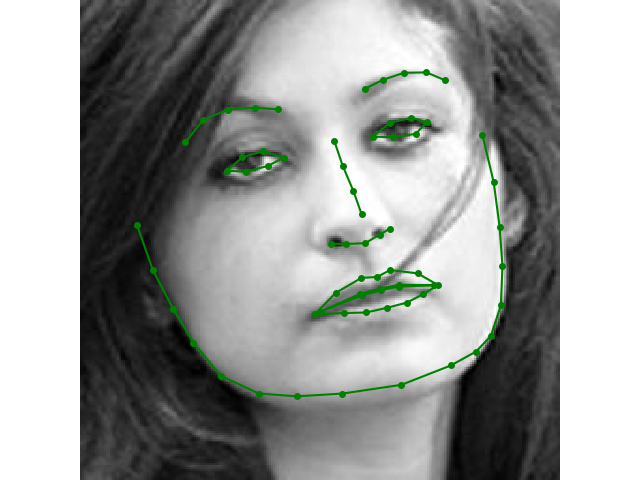}
    \end{minipage}
    %\begin{minipage}[t]{0.11\textwidth}
        %\centering
        %\includegraphics[width=2.4cm]{353_sparse.png}
    %\end{minipage}
    \begin{minipage}[t]{0.13\textwidth}
        \centering
        \includegraphics[width=2.7cm]{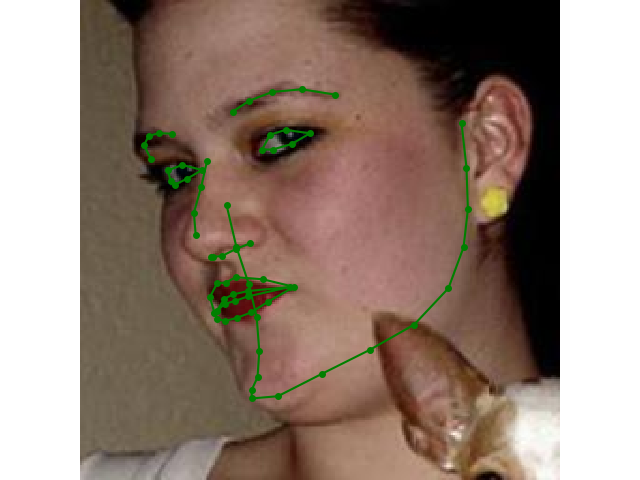}
    \end{minipage}
    %\begin{minipage}[t]{0.11\textwidth}
        %\centering
        %\includegraphics[width=2.4cm]{552_sparse.png}
    %\end{minipage}
    %\begin{minipage}[t]{0.11\textwidth}
        %\centering
        %\includegraphics[width=2.4cm]{577_sparse.png}
    %\end{minipage}
    %\begin{minipage}[t]{0.11\textwidth}
        %\centering
        %\includegraphics[width=2.4cm]{798_sparse.png}
    %\end{minipage}
    \begin{minipage}[t]{0.13\textwidth}
        \centering
        \includegraphics[width=2.7cm]{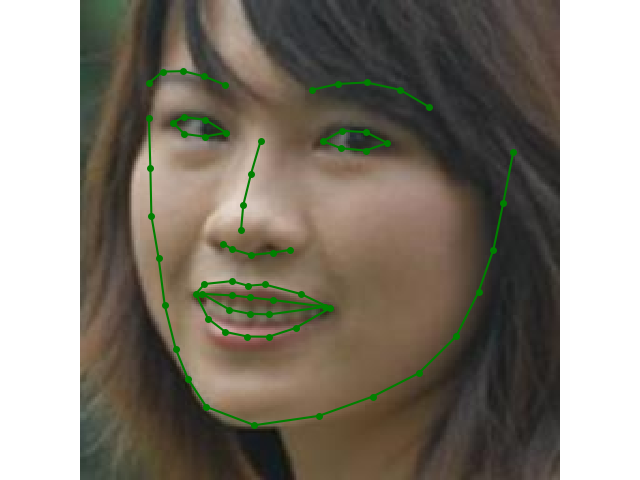}
    \end{minipage}
    \begin{minipage}[t]{0.13\textwidth}
        \centering
        \includegraphics[width=2.7cm]{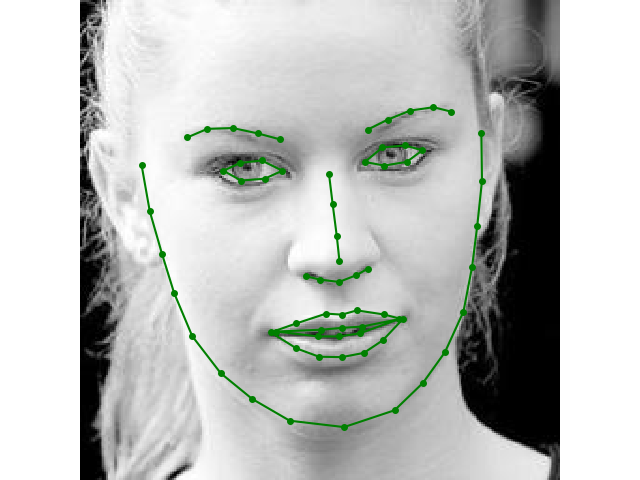}
    \end{minipage}
    \begin{minipage}[t]{0.13\textwidth}
        \centering
        \includegraphics[width=2.7cm]{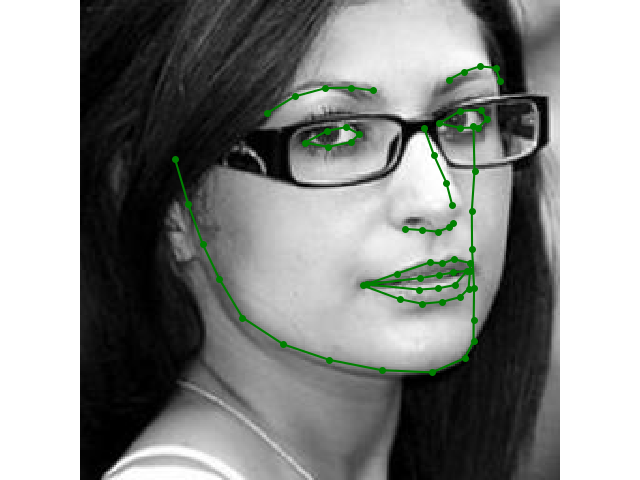}
    \end{minipage}
    %\begin{minipage}[t]{0.13\textwidth}
        %\centering
       % \includegraphics[width=2.7cm]{RESULTS/1427_sparse.png}
   % \end{minipage}
    %\begin{minipage}[t]{0.13\textwidth}
        %\centering
        %\includegraphics[width=2.7cm]{969_sparse.png}
%    \end{minipage}

    \begin{minipage}[t]{0.13\textwidth}
        \centering
        \includegraphics[width=2.7cm]{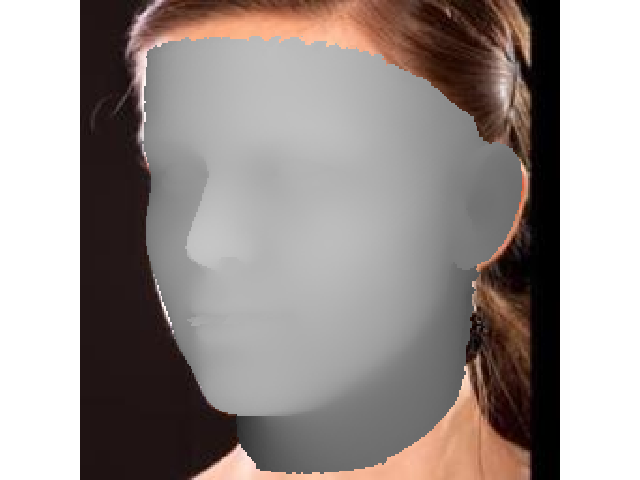}
    \end{minipage}
    \begin{minipage}[t]{0.13\textwidth}
        \centering
        \includegraphics[width=2.7cm]{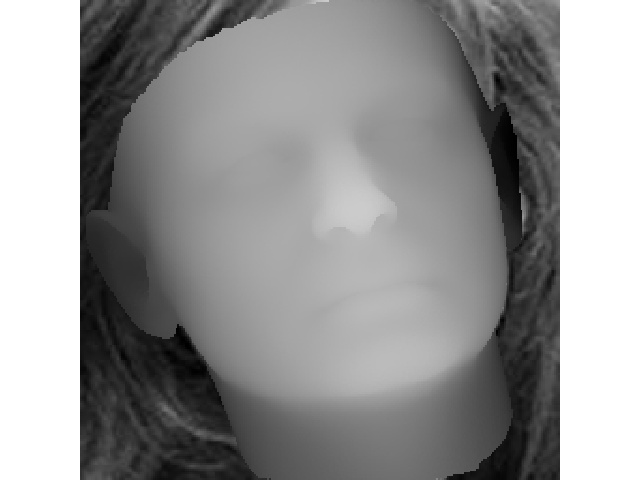}
    \end{minipage}
    %\begin{minipage}[t]{0.11\textwidth}
        %\centering
        %\includegraphics[width=2.4cm]{353_depth.png}
    %\end{minipage}
    \begin{minipage}[t]{0.13\textwidth}
        \centering
        \includegraphics[width=2.7cm]{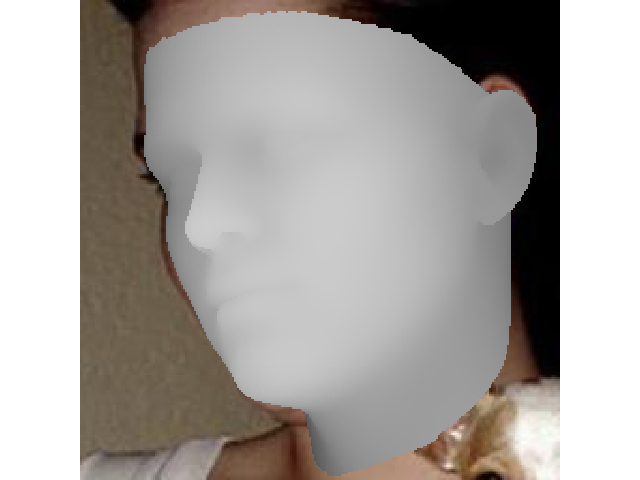}
    \end{minipage}
    %\begin{minipage}[t]{0.11\textwidth}
        %\centering
        %\includegraphics[width=2.4cm]{552_depth.png}
    %\end{minipage}
    %\begin{minipage}[t]{0.11\textwidth}
        %\centering
        %\includegraphics[width=2.4cm]{577_depth.png}
    %\end{minipage}
    %\begin{minipage}[t]{0.11\textwidth}
        %\centering
        %\includegraphics[width=2.4cm]{798_depth.png}
    %\end{minipage}
    \begin{minipage}[t]{0.13\textwidth}
        \centering
        \includegraphics[width=2.7cm]{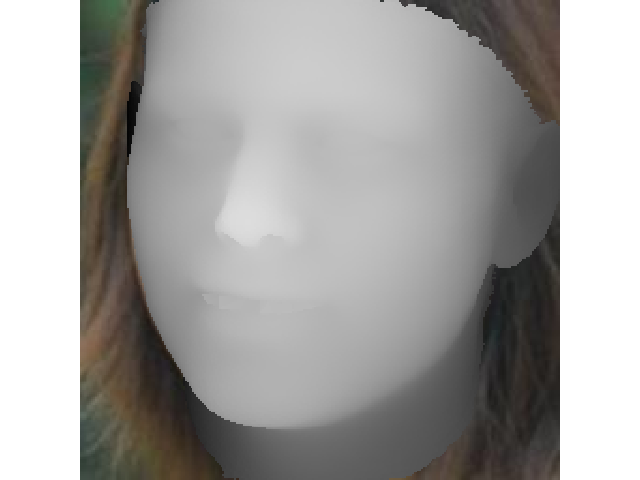}
    \end{minipage}
    \begin{minipage}[t]{0.13\textwidth}
        \centering
        \includegraphics[width=2.7cm]{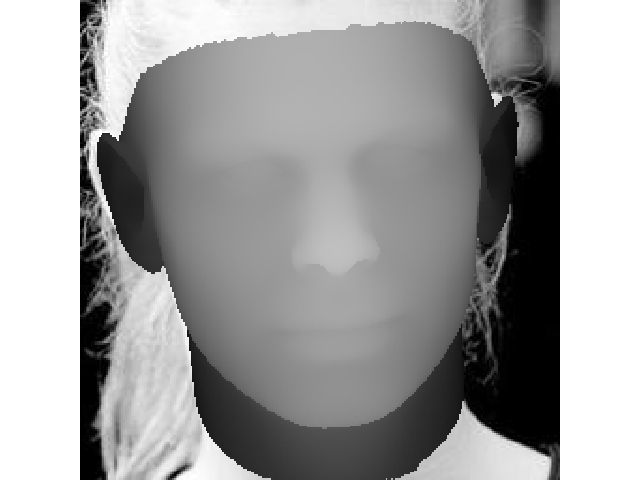}
    \end{minipage}
    \begin{minipage}[t]{0.13\textwidth}
        \centering
        \includegraphics[width=2.7cm]{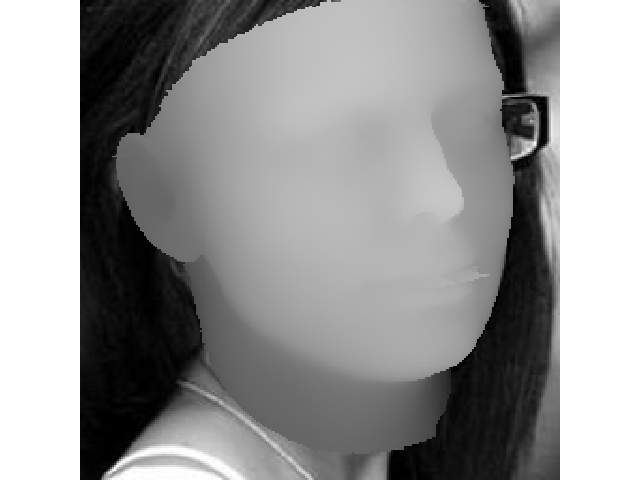}
    \end{minipage}    
    %\begin{minipage}[t]{0.13\textwidth}
       % \centering
      %  \includegraphics[width=2.7cm]{RESULTS/1427_depth.png}
  %  \end{minipage}
    %\begin{minipage}[t]{0.11\textwidth}
        %\centering
        %\includegraphics[width=2.7cm]{969_depth.png}
    %\end{minipage}
    
    \begin{minipage}[t]{0.13\textwidth}
        \centering
        \includegraphics[width=2.7cm]{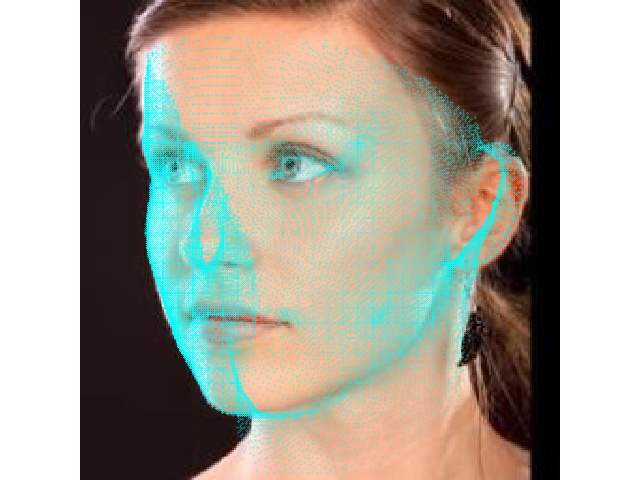}
    \end{minipage}
    \begin{minipage}[t]{0.13\textwidth}
        \centering
        \includegraphics[width=2.7cm]{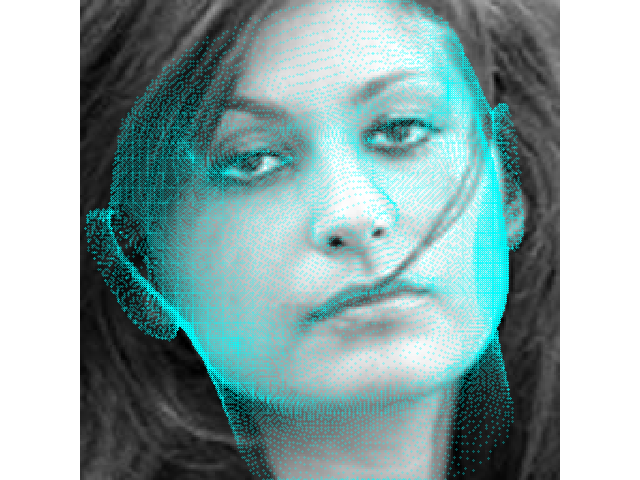}
    \end{minipage}
    %\begin{minipage}[t]{0.11\textwidth}
        %\centering
        %\includegraphics[width=2.4cm]{353_dense.png}
    %\end{minipage}
    \begin{minipage}[t]{0.13\textwidth}
        \centering
        \includegraphics[width=2.7cm]{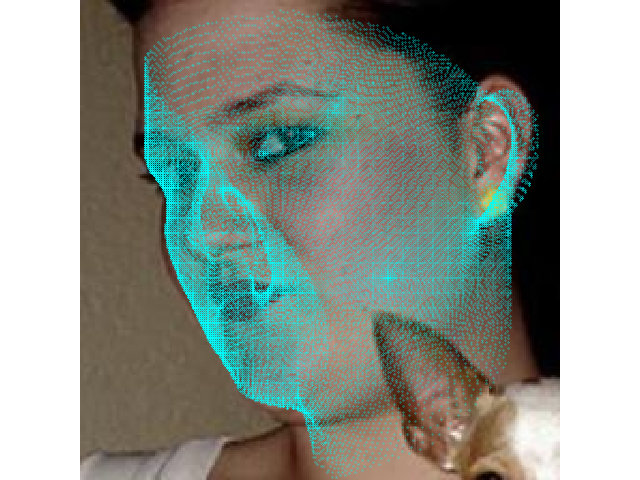}
    \end{minipage}
    %\begin{minipage}[t]{0.11\textwidth}
        %\centering
        %\includegraphics[width=2.4cm]{552_dense.png}
    %\end{minipage}
    %\begin{minipage}[t]{0.11\textwidth}
        %\centering
        %\includegraphics[width=2.4cm]{577_dense.png}
    %\end{minipage}
    %\begin{minipage}[t]{0.11\textwidth}
        %\centering
        %\includegraphics[width=2.4cm]{798_dense.png}
    %\end{minipage}
    \begin{minipage}[t]{0.13\textwidth}
        \centering
        \includegraphics[width=2.7cm]{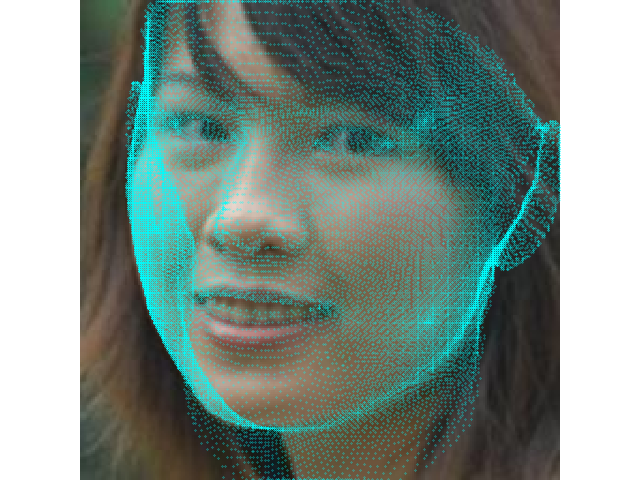}
    \end{minipage}
    \begin{minipage}[t]{0.13\textwidth}
        \centering
        \includegraphics[width=2.7cm]{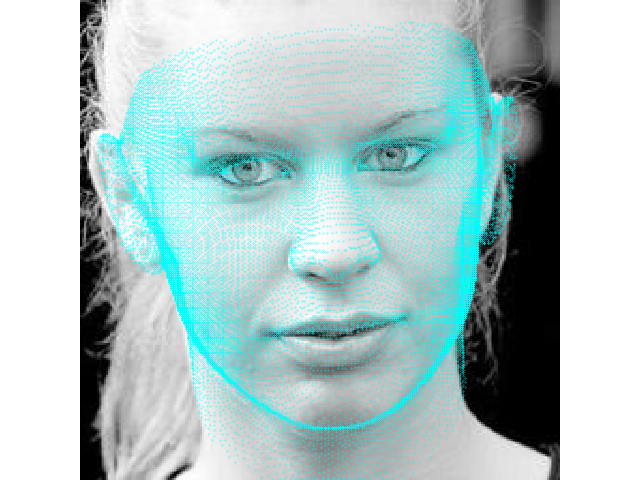}
    \end{minipage}
    \begin{minipage}[t]{0.13\textwidth}
        \centering
        \includegraphics[width=2.7cm]{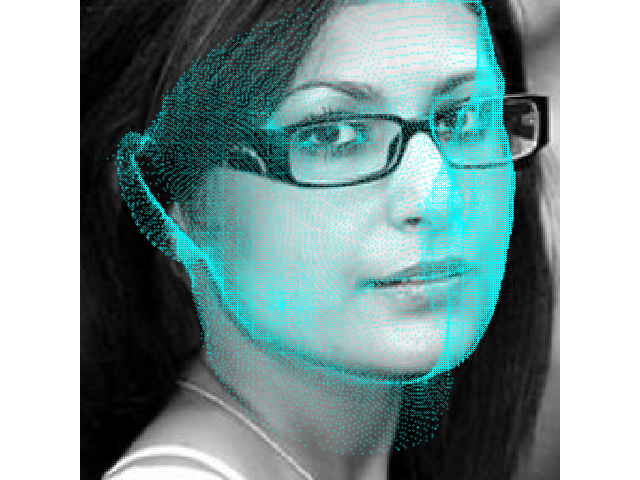}
    \end{minipage}
    \caption{Qualitative results of face alignment on AFLW2000-3D dataset  \cite{zhu2016face}. Top row: Sparse face alignment results with 68 landmarks plotted, including eyes, eyebrows, nose, mouth, and jawline. Middle row: Faces rendered with the reconstructed depth map. Bottom row: Dense face alignment results with all the 53,215 landmarks plotted. Note, although the results are good as shown by these faces in front view, it may seem the overlays dislocated for faces of side views because the reconstruction is only for the front view as the ground truth available for training is front view. } 
    \label{fig5}
\end{figure*}

\subsection{Most Recent Work}%\vspace{-0.1em}
On the basis of  \cite{ranjan2018generating},  \cite{cheng2019meshgan} proposed an intrinsic adversarial architecture to reconstruct more detailed 3D face mesh, and  \cite{zhou2019dense} reconstructed the 3D face mesh from a 2D image, in particular, CMD  \cite{zhou2019dense} added additional texture information in the graph structure to simultaneously regress coordinates and colour of the mesh. However, \cite{zhou2019dense} used encoder-decoder networks to reconstruct the 3D mesh, and our work is different from them. As they all utilize encoder to encode the 2D image into latent embeddings with CNNs and decoder that reconstructs the 3D face mesh with GCNs from the latent embeddings. It is believed that only using latent embeddings to represent 2D information is not enough, as some low-level semantic features cannot be represented properly and feature information will be lost during the down-sampling or encoding process. Furthermore, the same situation happens in the up-sampling process. The decoder cannot recover the lost resolution and semantic information very well when latent embeddings are the only input information. 

For the 3D face problem, overall facial structure is fixed, semantic information is not very rich, so the low-level semantic information and high-level spatial features may both be valuable. We propose a multi-level regression mappings mechanism between each down-sampled 2D image feature and corresponding 3D face mesh features, equipped with a resolution preserved and feature aggregated network structure. We focus on fusing different depth features along different paths in networks. Our proposed method gains superior performance on 2D and 3D face alignment tasks, especially in the large pose face alignment problem, because our model gains more useful information from visible parts of input face images, which help GCNs to better regressing the invisible mesh vertices.

\section{Method}%\vspace{-0.2em}
\subsection{Data Representation}%\vspace{-0.1em}

We represent the 3D face mesh with vertices and edges, \(F = (V,A)\) where V has N vertices in 3D Euclidean space, \(V\in \mathbb{R}^{N \times 3}\), and \(A\in\{0,1\}^{N \times N}\) is a a sparse adjacency matrix, representing the edge connections between vertices, where \(A_{i,j} = 1\) means vertices $V_i$,$V_j$ are connected by an edge, and \(A_{i,j} = 0\) otherwise. 

\subsection{Graph Fourier Transform}%\vspace{-0.1em}
Following  \cite{chung1997spectral}, the non-normalized graph Laplacian is defined as \(L = D - A \in \mathbb{R}^{N \times N}\), with \(D\) a diagonal matrix representing the degree of each vertex in \(V\), such that \(D_{i,i} = \sum_{j = 1}^{N} A_{i,j}\). The Laplacian of the graph is a symmetric and positive semi-definite matrix, so \(L\) can be diagonalized by the Fourier basis \(U\in{\mathbb{R}}^{N \times N}\), that \(L = U \Lambda U^{T} \). The columns of \(U\) are the orthogonal eigenvectors \(U = [u1,...,un]\), and \(\Lambda = diag([\lambda_{1},...,\lambda_{n}]) \in\mathbb{R} ^{N \times N}\) is a diagonal matrix with real, non-negative eigenvalues. The graph Fourier transform of the face representation \(x \in \mathbb{R} ^{N \times 3}\) is defined as \(\hat{x} = U ^{T} x\), and the inverse Fourier transform as \(x = U \hat{x}\).

\subsection{Spectral Graph Convolution}%\vspace{-0.1em}
The convolution operation on a graph can be defined in Fourier space by formulating mesh filtering with a kernel \(g_{\theta}\) using a recursive Chebyshev polynomial  \cite{defferrard2016convolutional}. The filter \(g_{\theta}\) is parametrized as a Chebyshev polynomial expansion of order \(K\) such that

\begin{equation}
	g_{\theta}(L) =  \sum_{k =1}^{K} \theta_{k} T_{k} (\hat{L} )
\end{equation}

where \(\hat{L} = 2L/\lambda_{max} - I_{N}\) represents rescaled Laplacian, and parameter \(\theta_{k}\) is a vector of Chebyshev coefficients. \(T_{k} \in\mathbb{R} ^{N \times N}\) is the Chebyshev polynomial of order \(K\), that can be recursively computed as \(T_{k} (x) = 2xT_{k-1}(x) - T_{k-2}(x) \) with \(T_{0} = 1\) and \(T_{1} = x\). Therefore, the spectral convolution can be defined as

\begin{equation}
	 y_{j} = \sum_{i =1}^{F_{in}} g_{\theta_{i,j}} (L)x_{i} 
\end{equation}
where input \(x \in \mathbb{R} ^{N \times F_{in}}\) has \(F_{in}=3\) features, as the face mesh of vertices is 3D and \(y\in\mathbb{R} ^{N \times{F_{out}}}\) is the output. This approach is computationally faster and complexity drops from $\mathcal{O}(n^{2}) $ to $\mathcal{O}(n)$, compared with  \cite{bruna2013spectral}.

\begin{figure*}[t] %pic result
	\centering
	%\begin{subfigure}%{0.14\textwidth}
		%\centering
		%\includegraphics[width=1.6cm]{operation_order.png}
	%	\captionsetup{font={scriptsize}}
	%\end{subfigure}
% 	\begin{subfigure}%{0.6\textwidth}
% 		\centering
	\includegraphics[width=18cm]{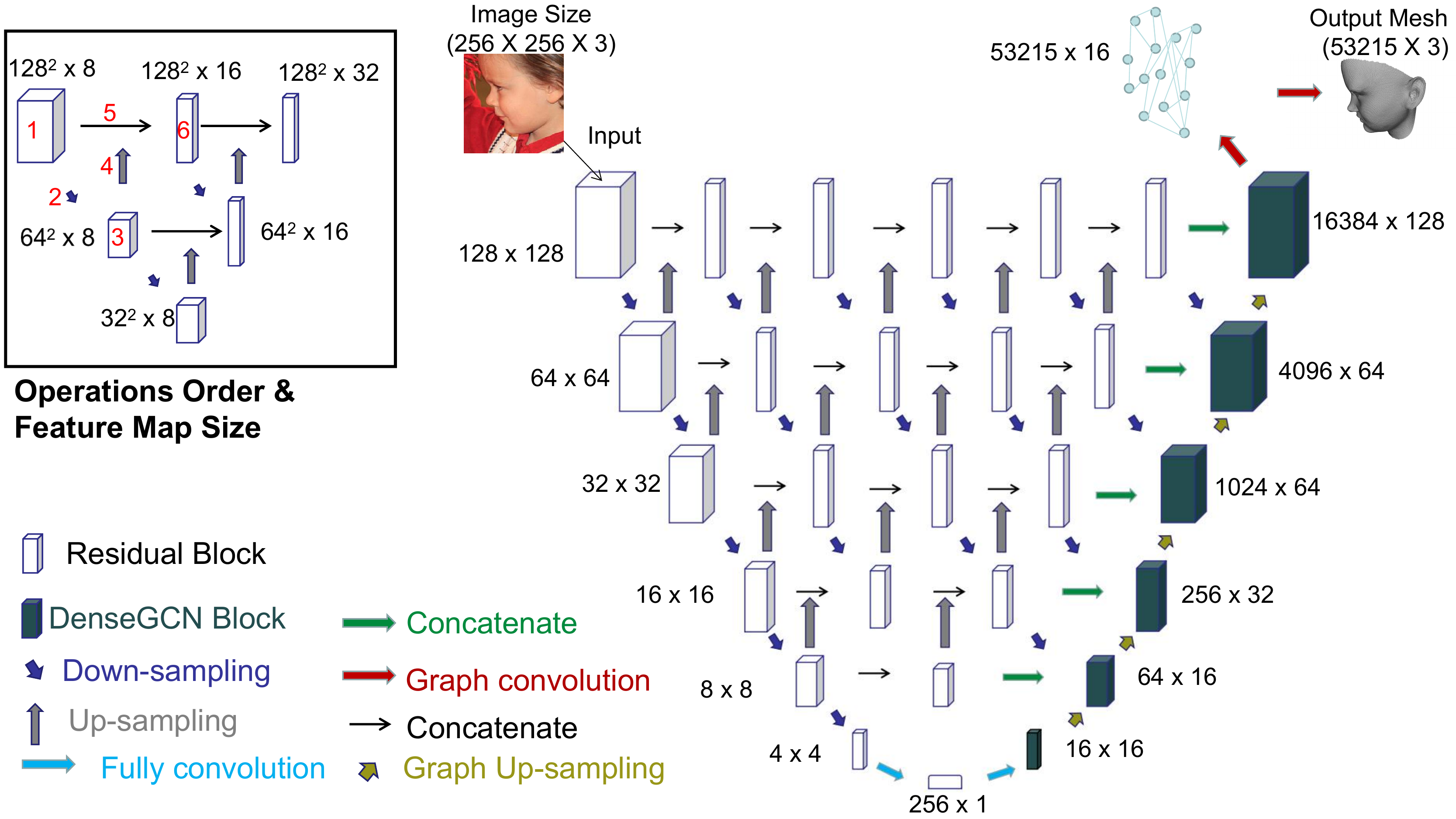}
		%\captionsetup{font={scriptsize}}
% 	\end{subfigure}
	%\begin{subfigure}%{0.18\textwidth}
		%\centering
		%\includegraphics[width=2.2cm]{legend_02.png}
		%\captionsetup{font={scriptsize}}
	%\end{subfigure}
	\caption{Overview of our proposed model. Down-sampling is conducted by setting stride size in the convolution layers as 2. Lower level features are bilinearly up-sampled by a factor 2. On the left branch, we show the feature map size after down-sampling, and on the right branch, we show the vertex feature map size with channels after up-sampling, because we use a vector to represent each vertex. For example, 16384$\times$128 means that 16384 vertices are maintained, and each vertex is represented by a 128$\times$1 vector. The order of operations and feature map size in a small level of aggregate circulation are illustrated in the left side, following the ascending order from 1 to 6 (in red color).  As is shown, number 5 is the concatenation of number 1 and number 4's output, then as input to number 6. The green arrow concatenates the output from CNN Residual Block and DenseGCN Block at the same level. Graph down-sampling process is not shown because of the space limitation. More details can be found in Section 3.4.}
    \label{3}
\end{figure*}

\subsection{Mesh Sampling}%\vspace{-0.1em}
To achieve multi-scale graph convolutions on joint mesh vertices and 2D feature maps from CNNs, we follow  \cite{ranjan2018generating} to form a new topology and neighbour relationships of vertices. More specifically, we use the permutation matrix \(Q_{d} \in\{0,1\}^{m \times{n}}\) to down-sample a mesh with \(m\) vertices. \(Q_{d}(i,j) = 1\) denotes the \(j^{th}\) vertex is kept, and \(Q_{d}(i,j) = 0\) otherwise. Up-sampling is conducted with another transformation matrix \(Q_{u} \in\mathbb{R} ^{m \times n}\). In order to train the CNN and GCN hierarchically and iteratively, we specially design the number of vertices that are maintained in each up-sampling stage, and the feature map size in the down-sampling process, to enable CNNs and GCNs to cooperate in the same level. More details will be shown in Section 3.5.

Down-sampling in GCN is obtained by iteratively contracting vertex pairs, which uses a quadratic matrix to maintain surface error approximations  \cite{garland1997surface}. The discarded vertices during down-sampling are recorded using barycentric coordinates. The up-sampling operates convolution transformations on retained vertices and map the discarded vertices into the down-sampled mesh surface using Barycentric coordinates. The up-sampled mesh with vertices \(V_{u}\) is obtained by a sparse matrix multiplication, i.e., \(V_{u} = Q_{u}V_{d}\), where \(V_{d}\) are down-sampled vertices.

\subsection{Proposed Aggregation Network}%\vspace{-0.1em}% can GCN be deeper?  U-net for generation
Our novel aggregation graph regression network is motivated by fusing features hierarchically and iteratively \cite{yu2018deep,zhou2018unet++}, which is illustrated in Fig. \ref{3}.   Our model can provide improvements in extracting the full spectrum of semantic and spatial information across stages and resolutions. Our network consists of an encoder and a decoder, which are connected by a series of nested residual convolution blocks (aggregation block). As we mentioned in section 3.4, we specially design the number of vertices that remain after each up-sampling stage in the decoder, and the feature map size after each convolution block in the encoder to make them equal with each other. For example, after first Residual Block, we make the feature map size 128$\times$128, which is equal to the total number of vertices remained after the last DenseGCN block 16384. Our experiments show that maintaining the same graph nodes on the same level helps to improve the performance in our model. We achieve direct end-to-end regression from 2D image to 3D mesh vertices through different feature levels, by making CNNs cooperate with GCNs directly.

The encoder takes input images of shape 256$\times$256$\times$3, and has six residual convolution blocks  \cite{he2016deep}. After each residual convolution block, the feature map size is decreased by half. This reduction continues until the dimension becomes 4$\times$4$\times$128. Then two fully connected layers are applied to construct a 256$\times$1 dimension embedding.

An aggregation block contains a series of residual convolution blocks, in which there are three convolution layers with identity short-cut connection followed by a Batch Normalization layer  \cite{ioffe2015batch} and Leaky Relu as the activation function. For each filter, the kernel size is three and the stride is one. Different from the network proposed by  \cite{yu2018deep}, our aggregation block achieves fully fused local and global information from encoder into the decoder. Up-sampling operations in the aggregation block from shallow to in-depth, further refining features when extracting 2D image features. Besides, we add down-sampling operations which can project high-resolution features from 2D images into low-resolution 3D mesh features. With up-sampling and down-sampling operations, the aggregation block can extract and reuse more features through different resolutions and scales, which can help to decrease information loss during the encoding process. In Section 5.3, our ablation study demonstrates that the combination of up-sampling and down-sampling helps to extract more useful information. Finally, the aggregation block iteratively and hierarchically aggregates these operations to learn a deep fusion of low and high-level feature information.

The decoder takes embeddings and multi-level outputs from the aggregation block, then decodes with six dense graph convolution blocks (DenseGCN), inspired by  \cite{li2019can}. It has been shown that as layers go deeper, DenseGCN can prevent vanishing gradient problems. Our DenseGCN block consists of 4 graph convolution layers, and each graph convolution layer is followed by a Batch Normalization layer  \cite{ioffe2015batch} and Leaky Relu. After 6 DenseGCN blocks and graph up-sampling operations, the number of vertices is up-sampled from 16 to 16384, and each vertex is represented by a vector of length 128. At last, two graph convolution layers are added to generate a 3D face mesh, which up-samples the number of vertices to 53215 and reduces the vertex feature map channels to 3, as each face mesh vertex has three dimensions: x, y, and z. On the right branch of the network structure in Fig. \ref{3}, we show the process of up-sampling vertices hierarchically with face meshes. Through an ablation study in Section 5.3, we demonstrate that our proposed method can perform better than non-aggregation or shallow aggregation network in the 3D face alignment task.

\subsection{Loss Function}%\vspace{-0.1em}
L2 and L1 loss have widely been used in facial landmark localization tasks by CNN based networks. It is commonly known that the L2 loss is sensitive to outliers, so in the early training stage, the training process can be unstable. With the L1 loss, it is difficult to continuously converge and find the global minimization in the late training stage without careful tuning of the learning rate. Most of the facial landmarks localization methods use a joint loss function to guide the training process. For example, PRN  \cite{feng2018joint} uses a weighted L2 loss function to make the model pay more attention to the central region of the face. CMD \cite{zhou2019dense} uses a joint loss function where the L2 loss for shape reconstruction, L1 for texture regression and L-render to minimize pixel-wise reconstruction error for facial pixels rendering. %Furthermore, \cite{wei20193d} also utilizes a joint L1 and L-smooth loss function to make the generated face surface more smooth, where L-smooth helps to decrease the difference between each vertex and its surrounding vertices.

Inspired by Wing-loss  \cite{feng2018wing} and Smooth-L1 loss in  \cite{girshick2015fast}, we propose a new loss function that can prevent the model from taking large update steps when approaching small range errors in the late training stage and can recover quickly when dealing with large errors during the early training stage. Our loss function is defined as:

\begin{equation}
L(x) = 
\left\{
\begin{array}{lr}
W [e^{(|x|/\epsilon)} - 1] & if |x| < W  \\
|x| - C & otherwise \\

\end{array}
\right.
\end{equation}
Where $W$ should be non-negative and limit the range of the non-linear part, $\epsilon$ decides the curvature between \((-W,W)\) and $ C = W - W[e^{(|w|/\epsilon)} - 1] $ connects the linear and non-linear parts. After several evaluation experiments, the parameter \(W\) is set to 5 and \(\epsilon\) to 4 in this work.
\section{Experiments}%\vspace{-0.1em}

\subsection{Datasets}%\vspace{-0.1em}
We train our model using semi-annotated in-the-wild data (300W-LP)  \cite{zhu2016face}. The 300W-LP dataset contains \(61225\) large pose facial images with corresponding 3DMM parameters and pose coefficients, which are synthetically generated by the profiling method  \cite{zhu2016face}. The dataset is produced by fitting a 3DMM model using the multi-feature fitting approach (MFF)  \cite{romdhani2005estimating}. Each image is rendered to 10-15 different poses resulting in a large scale dataset.

For the evaluation of the trained model, we perform extensive quantitative experiments on AFLW2000-3D  \cite{zhu2016face} dataset. It contains 2000 large pose samples from the AFLW dataset  \cite{koestinger2011annotated}, annotated with fitted 3DMM parameters and 68 3D landmarks. The sparse and dense face alignment evaluations are performed on this dataset. 

AFLW-LFPA is another extension of AFLW dataset constructed by \cite{jourabloo2016large}. According to the poses, the dataset contains 1299 test images with a balanced distribution of yaw angles. Besides, each image is annotated with 13 additional landmarks as a expansion to the original 21 visible landmarks in AFLW. Same as \cite{feng2018joint}, We use 34 visible landmarks as the ground truth to measure the accuracy of our results. This database is evaluated on the task of sparse 3D face alignment. 

The Florence dataset is a 3D face dataset that contains high-resolution 3D scans of 53 samples which are acquired from a structure-light scanning system. We compare the performance of our method on face reconstruction against other recent state-of-the-art methods.

\subsection{Implementation Details}%\vspace{-0.1em}
We first fit the Basel Face Model (BFM)  \cite{blanz1999morphable} model to generate and transform the 3D face mesh with corresponding pose coefficients to form the training data set. Specifically, we crop the images according to the ground truth bounding box and rescale them into size 256$\times$256. To augment our dataset, similar to other methods  \cite{feng2018joint}, we perturb the input image by randomly rotating and translating. Specifically, the rotation ranges from $-45$ to $45$ degree angles, translation changes is random from 10\% of the input image size and has a scale range from 0.9 to 1.2. We use stochastic gradient descent with a momentum of 0.9 to optimize our loss function. We trained our model with a learning rate of 1e-3 and decay rate of 0.99 every epoch.  The order of Chebyshev polynomial is set to 3 for all the graph convolution layers. The batch size is set as 48. All training processes are performed on a server with 8 TESLA V100, and all test experiments are conducted on a local machine Geforce RTX 2080Ti.
%All the experiments are conducted on a Geforce RTX 2080Ti (NVIDIA, Santa Clara, CA).
\begin{table}
\centering
\caption{Face alignment results on AFLW2000-3D benchmarks. The performance is reported as bounding box size normalized mean error (\%). The best result in each category is highlighted in bold, the lower value is better. For any specific head pose, our model outperforms the other methods, and in particular, it defeats the other methods by a large margin for large pose yaw ($60^\circ$ to $90^\circ$).}
\scalebox{0.8}{
\begin{tabular}{c||c|c|c|c||c}
    \hline
    \multicolumn{1}{c||}{\multirow{2}{*}{Methods}} & \multicolumn{4}{c||}{AFLW2000-3D} & \multicolumn{1}{c}{AFLW-LFPA} \\ \cline{2-6} 
\multicolumn{1}{c||}{} & \multicolumn{1}{c|}{$0^\circ\sim30^\circ$} & \multicolumn{1}{c|}{$30^\circ\sim60^\circ$} & \multicolumn{1}{c|}{$60^\circ\sim90^\circ$} & \multicolumn{1}{c||}{Mean} & \multicolumn{1}{c}{Mean} \\ \hline
    SDM \cite{xiong2015global}         & 3.67    & 4.94     & 9.67     & 6.12 & -        \\ 
    3DDFA \cite{zhu2016face}       & 3.78    & 4.54     & 7.93     & 5.42   &  -    \\ 
    3DDFA + SDM \cite{zhu2016face} & 3.43    & 4.24     & 7.17     & 4.94     &  -  \\ 
    N3DMM \cite{tran2018nonlinear}  & -        & -        & -        & 4.70	&	-	\\ 
    DeFA \cite{liu2017dense}        & -       & -        & -        & 4.50        & 3.86 \\ 
    3DSTN \cite{bhagavatula2017faster}       & 3.15    & 4.33     & 5.98     & 4.49       &  -\\
    CMD \cite{zhou2019dense}			& -       & - 		& -			& 3.98 & -	\\
    PRN \cite{feng2018joint}         & 2.75    & 3.51     & 4.61     & 3.62       &  2.93 \\ 
    %Huawei \textit{et al.} \cite{wei20193d}  		& 2.44 	  & 3.26 	& 4.35		&3.35 & 2.65	\\
    Bulat \textit{et al.} \cite{bulat2017binarized} &2.47 &\textbf{3.01} &4.31 &3.26 & -  \\
    Jia \textit{et al.} \cite{guo2018stacked} & - & - & - & 3.07 & -  \\  \hline
    Ours     & \textbf{2.38} & 3.03 &\textbf{3.54} & \textbf{2.98} & \textbf{2.86} \\ \hline
\end{tabular}
}

\label{compar}
%\vspace{-3em}
\end{table}

\section{Results}%\vspace{-0.1em}
In this section, we show our qualitative and quantitative results on AFLW2000-3D  \cite{zhu2016face} and Florence  \cite{bagdanov2011florence} dataset in comparison with several other state-of-the-art methods. We then showed the results of an ablation study in order to demonstrate the effectiveness of the proposed aggregation block. The qualitative results of face alignment and 3D face reconstruction are shown in Fig. \ref{fig5} and Fig. \ref{Florence} (b) respectively.

\begin{figure}[H]
  \begin{minipage}{\linewidth}
    \centering
    \subcaptionbox{68 points with 2D coordinates.}
    {\includegraphics[height=3cm]{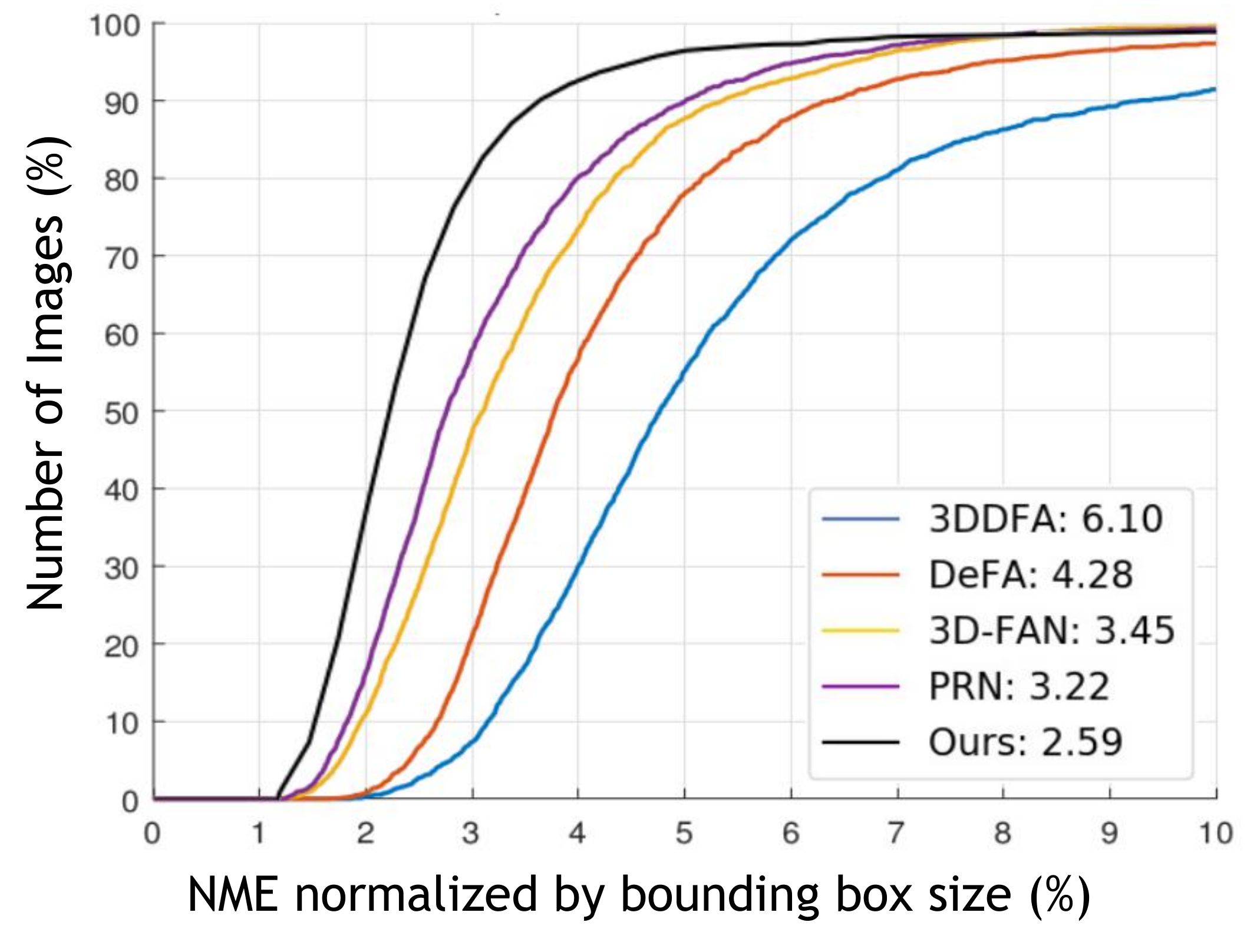}}\quad
    \subcaptionbox{68 points with 3D coordinates.}
     {\includegraphics[height=3cm]{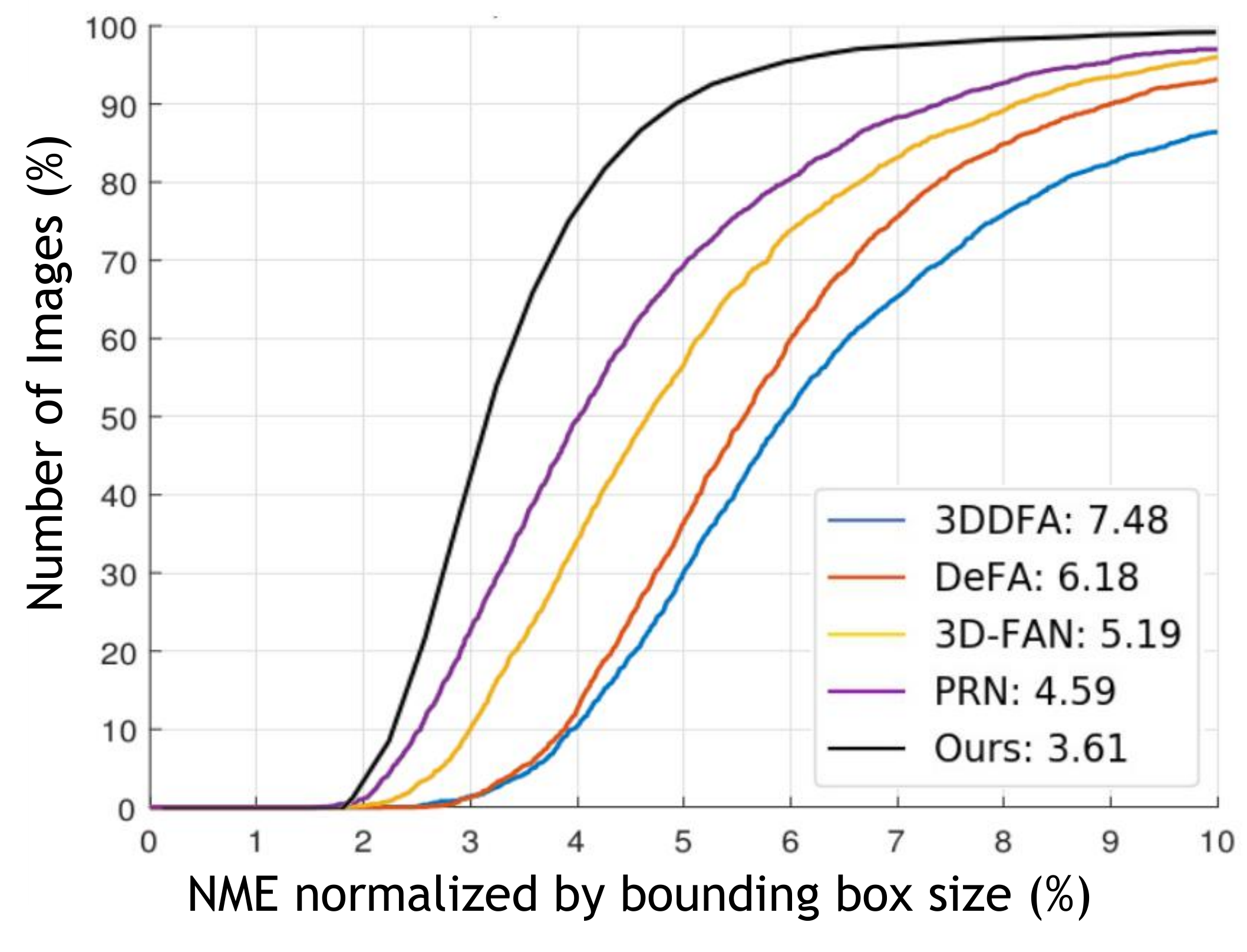}}
    \subcaptionbox{45K points with 2D coordinates.}
    {\includegraphics[height=3cm]{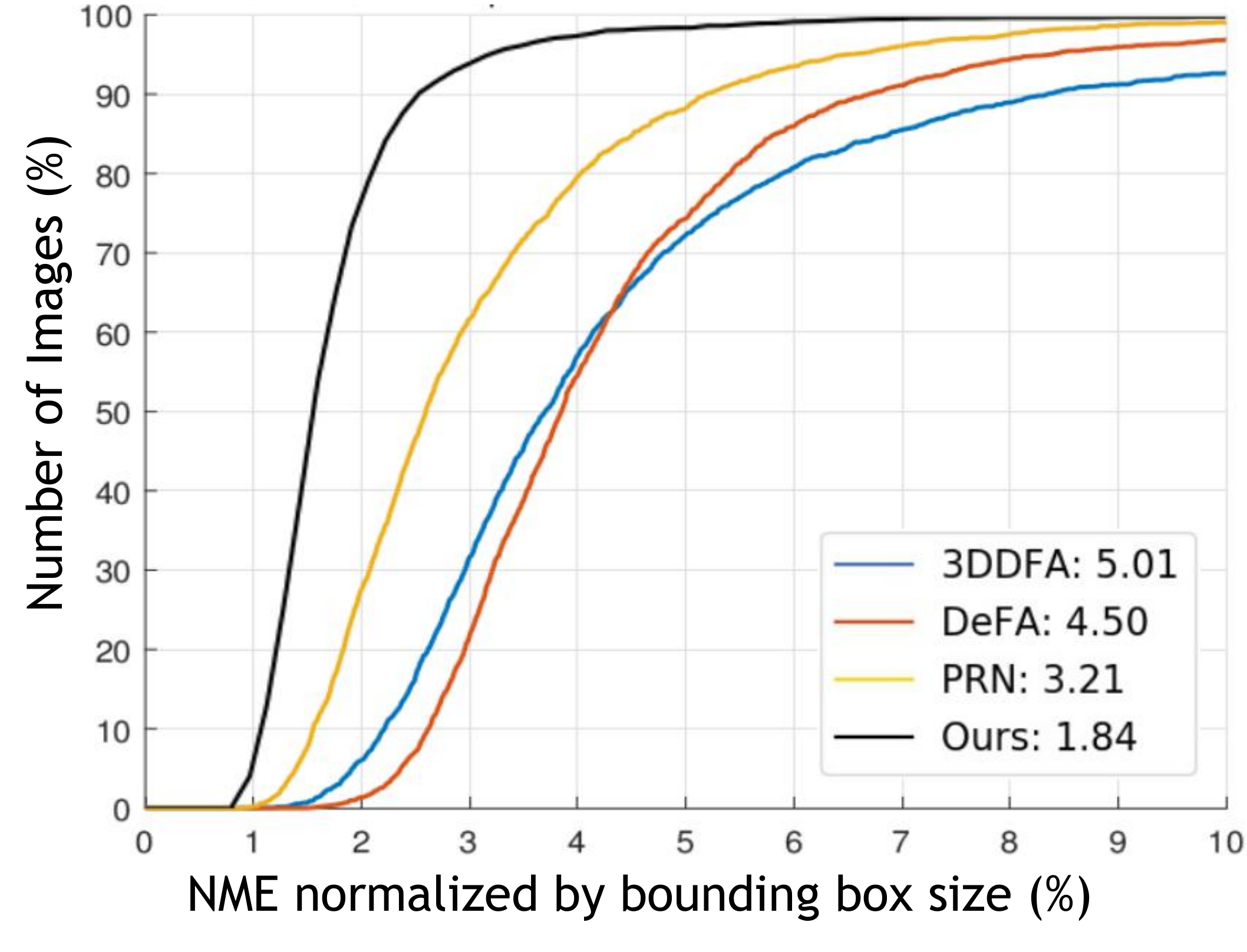}}\quad
    \subcaptionbox{45K points with 3D coordinates.}
    {\includegraphics[height=3cm]{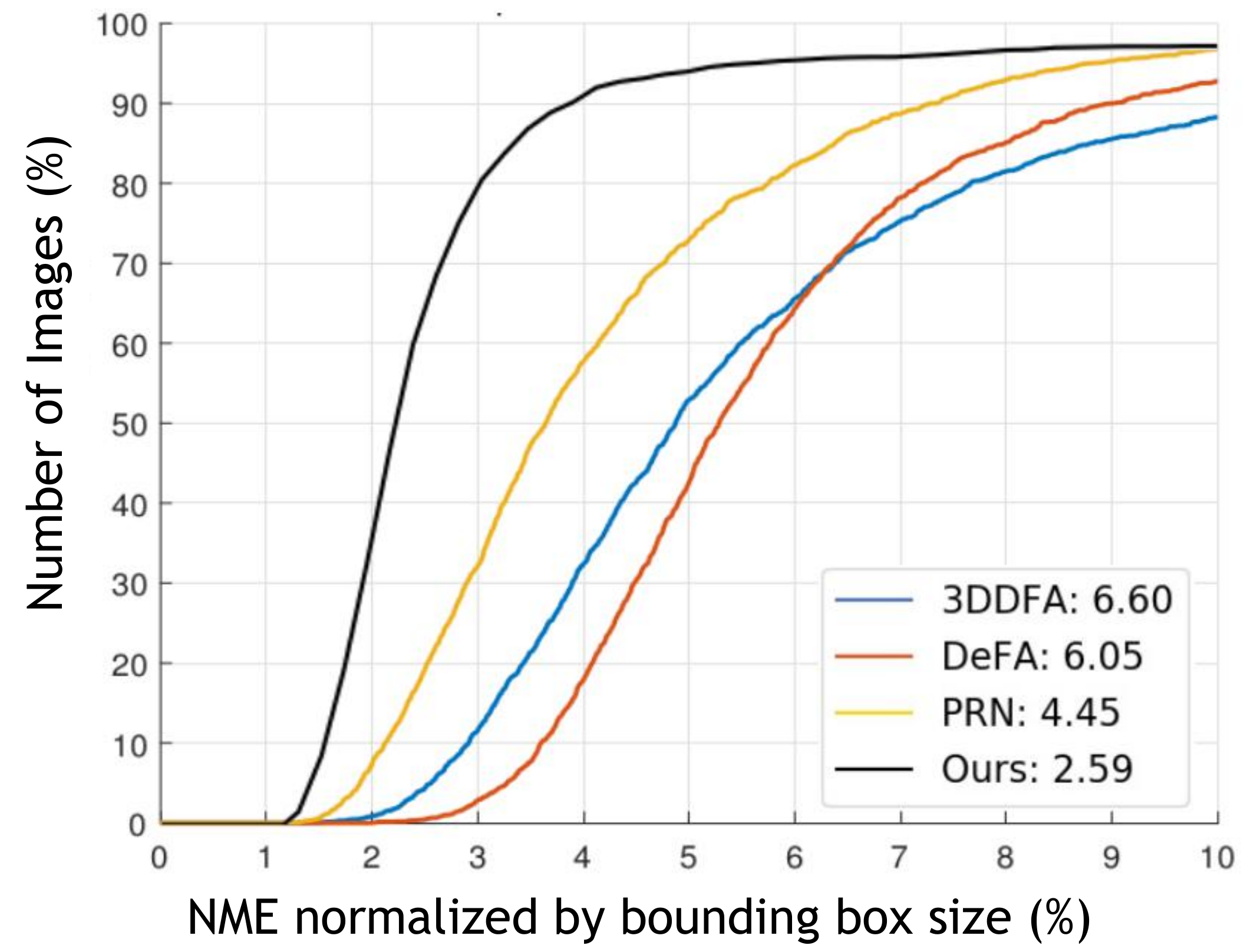}}\quad
    \caption{ Errors Distribution (CED) curves for sparse and dense face alignment on AFLW2000-3D. Note that for dense face alignment, PRN  \cite{feng2018joint} can only regress around 45K points, so we only select around 45K points for evaluation, even though our model can output all the 53215 vertices provided by the ground truth. Our model performs consistently better on both 2D and 3D problems when compared to other methods.}
    \label{9}
  \end{minipage}
\end{figure}

\begin{figure}[t] %pic result
	\centering
	\begin{minipage}[t]{0.23\textwidth}
		\centering
		\includegraphics[width=4cm]{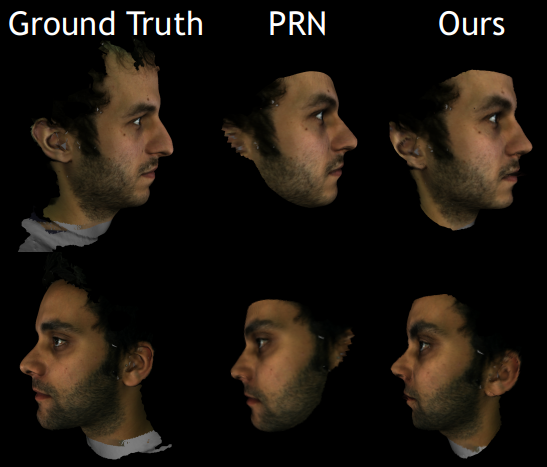}
		(a)
	\end{minipage}
	\begin{minipage}[t]{0.23\textwidth}
		\centering
		\includegraphics[width=4cm]{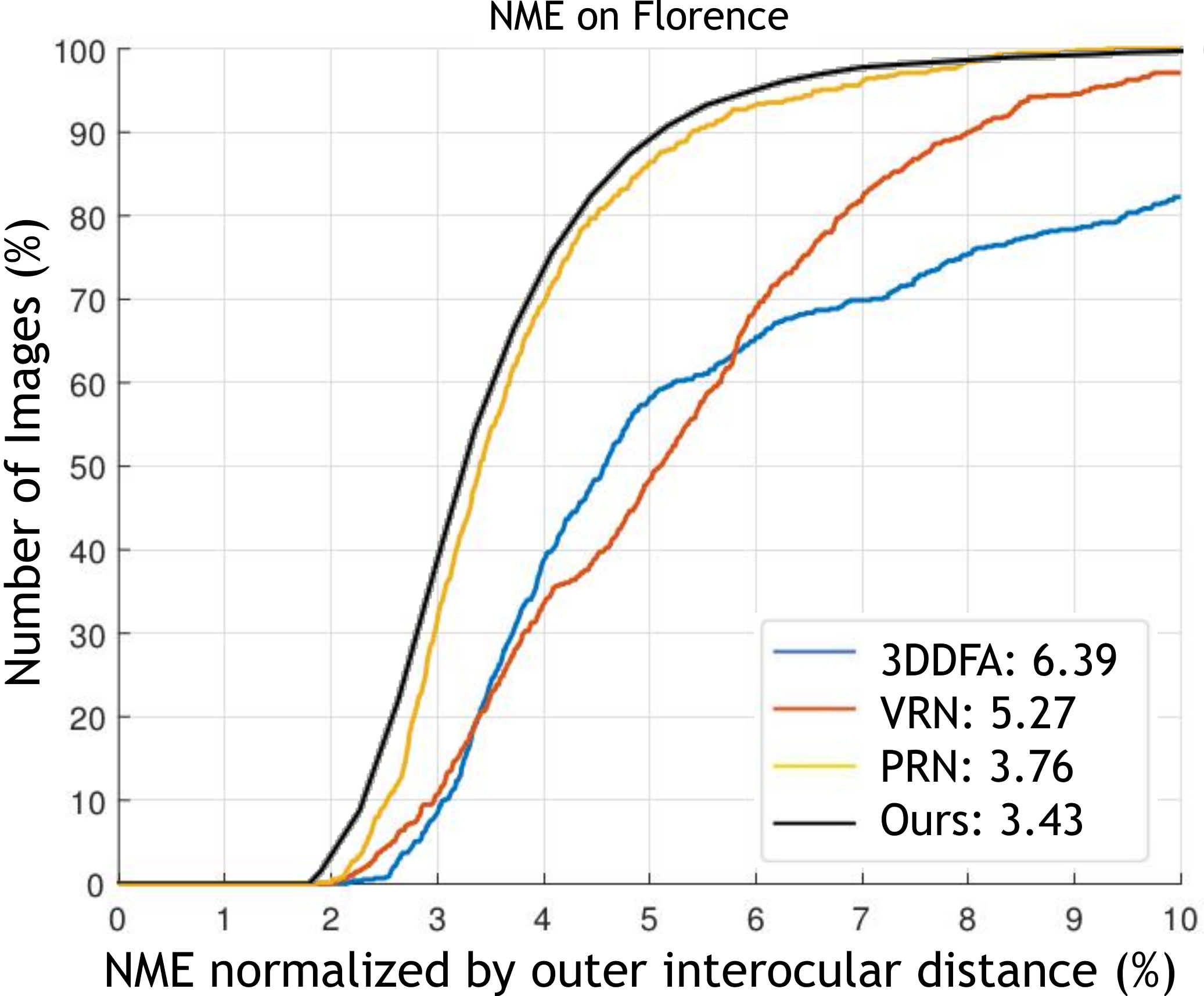}
		(b)
	\end{minipage}
	
    \caption{Example results on Florence dataset. (a): Qualitative results, First column are Ground truth  \cite{bagdanov2011florence}. The second column is Prediction by PRN  \cite{feng2018joint}. The third column is Results from our model. Note that our model can faithfully reconstruct more regions such as ears. (b): Quantitative results, the normalized mean error of each method is showed in the legend.}
    \label{Florence}
    %\vspace{-1em}
\end{figure}

\subsection{Face Alignment}%\vspace{-0.1em}
We compare our model with other state-of-the-art methods, 3DDFA \cite{zhu2016face}, DeFA  \cite{liu2017dense}, 3D-FAN  \cite{bulat2017far}, PRN  \cite{feng2018joint}, on sparse alignment tasks (68 landmarks). As suggested by 3DDFA \cite{zhu2016face}, normalized mean error (NME) is used as the alignment accuracy metric. NME is the average of the landmarks error normalized by the size of the bounding box. The bounding box size is defined as the rectangle hull of all the 68 landmarks, which is $\sqrt{width * height}$. Fig. \ref{9} (a) and (b) show the sparse face alignment with 68 landmarks on both 2D coordinate and 3D coordinate system. Our model exceeds other methods by a large margin on 3D face alignment. Specifically, more than 19 \% relative higher performance is achieved compared with the best method on both 2D and 3D coordinates.

Our model also produces good performance in a dense face alignment task with 45K vertices. We compare with previous state-of-the-art methods, including 3DDFA \cite{zhu2016face}, DeFA \cite{liu2017dense}, PRN \cite{feng2018joint}, and NME plots were shown in Fig. \ref{9} (c) and (d), which demonstrate that our model gains more than 41\% relative improvements compared to PRN \cite{feng2018joint}, so our model can produce more accurate vertices localization results, with the help of an aggregation block to extract more useful information and GCNs to directly perform feature learning on 3D face mesh.

We further evaluate our model on sparse face alignment with different face poses in 2D images in comparison with SDM \cite{xiong2015global}, 3DDFA \cite{zhu2016face}, 3DSTN \cite{bhagavatula2017faster}, DeFA \cite{liu2017dense}, PRN \cite{feng2018joint}, N3DMM \cite{tran2018nonlinear}, CMD \cite{zhou2019dense}. We randomly select 915 images from AFLW2000-3D to balance the distribution, whose absolute yaw angles with small, medium and large values are 1/3 each. Across three main classes with yaw values ($0^\circ\sim30^\circ$, $30^\circ\sim60^\circ$, $60^\circ\sim90^\circ$) for the faces in different images, our model exceeds the other state-of-the-art methods. Especially for large pose face alignment ($60^\circ\sim90^\circ$), as shown in Fig. \ref{fig5} our model can handle large pose face well. Because of the invisible parts of the face due to occlusion, the other methods cannot capture enough semantic information to regress the landmarks. Our model, however, utilizes aggregate feature learned from the visible part of the face to infer the unseen part of the faces' landmarks, which fuse and reuse the 2D semantic information to regress the 3D geometric information. The results are shown in Tab. \ref{compar}, where the numerical values of the other methods are cited from the original papers. As illustrated, our model achieves more than 25\% relative improvement over the best method on AFLW2000-3D dataset.

\subsection{3D Face Reconstruction}%\vspace{-0.1em}
We illustrate our model's ability in a 3D face reconstruction task with experiments on the Florence dataset  \cite{bagdanov2011florence}, compared with a state-of-the-art method, 3DDFA \cite{zhu2016face}, PRN \cite{feng2018joint}, VRN \cite{jackson2017large}, following the experimental settings in PRN \cite{feng2018joint}, and the metric which is the Mean Squared Error(MSE) normalized by outer interocular distance of 3D coordinates. We calculate the bounding box from the ground truth point cloud and crop the rendered image to 256$\times$256, and we follow \cite{jackson2017large} to choose 19K points of face region for evaluation. Note that, during the training process, our model only considers the coordinates of vertices, but for better visualization, the colors of faces are rendered from the corresponding input 2D image. For our model, we render colours to each 3D face vertex from the corresponding 2D input image pixels. Fig. \ref{Florence} shows the qualitative and quantitative results, our model can handle large pose face well and accurately covers more regions in lateral face parts, such as ears and necks, but PRN remains blurry in the ear area, and for quantitative comparison, our model achieves superior performance to PRN and outperform the other two methods by a large margin.
%and for invisible part of the 3D face in two dimensions plane, we just set their pixel value to zero.

\subsection{Ablation Study}%\vspace{-0.1em}
\noindent\textbf{Aggregation Block}:
In this section, we conduct several experiments to establish the effectiveness and compactness of our proposed aggregation block. We compare with no-aggregation block (encoder-decoder) and shallow-aggregation block structures (U-net). We change the decoder of those two networks into GCNs with graph up-sampling operations, but the encoder remains as CNNs. Apart from the aggregation part, the rest of the network maintain the same structure. Also, we remove the up-sampling and down-sampling operations, respectively, to further evaluate whether our aggregation block can help to better regress the face mesh vertices coordinates. Fig. \ref{12} (a) and (b) show the quantitative results on a 3D face alignment task. As is illustrated, our aggregation model attains a superior performance over the other four methods, and non-aggregation network structure has the worst performance. 

\noindent\textbf{Parameters of Loss Function}:
Several experiments are conducted to evaluate the parameter setting of our proposed loss function. Fig. \ref{12} (c) and (d) show the parameters setting results on sparse and dense alignments tasks, besides, our model is not sensitive to the two parameters, as no significant difference are found, and when w = 5, $\epsilon$ = 4, our model achieve best results.

\noindent\textbf{Loss Function}:
We compare with L1, L2, Smooth-L1 \cite{girshick2015fast} loss functions, which are commonly used in the regression problem. Experiments are performed on sparse alignment (68 points) and dense alignment (45K points) in 3D coordinates. The performance is reported as average NME(\%) of sparse and dense alignment tasks. Our proposed loss function (3.10 \%) outperforms smooth-L1 loss (3.39 \%) \cite{girshick2015fast} by 9 \% relatively better performance, L1 loss (3.61 \%) by 14 \% relatively better performance, and L2 Loss (4.02 \%) by 23 \% relatively better performance. Our proposed loss function attains a superior performance over the other three loss functions.

%Our method needs more time to converge, as more convolution operations are included during training.
\begin{figure}[H]
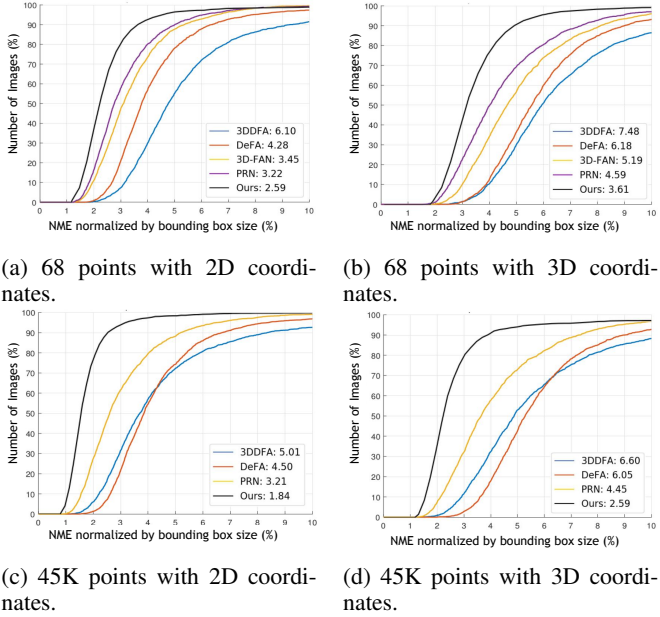

  \begin{minipage}{\linewidth}
    \centering
    \subcaptionbox{68 points with 2D coordinates.}
    {\includegraphics[height=3.2cm]{RESULTS/68_2.pdf}}\quad
    \subcaptionbox{68 points with 3D coordinates.}
     {\includegraphics[height=3.2cm]{RESULTS/68_3.pdf}}
    \subcaptionbox{45K points with 2D coordinates.}
    {\includegraphics[height=3.2cm]{RESULTS/53215_2.pdf}}\quad
    \subcaptionbox{45K points with 3D coordinates.}
    {\includegraphics[height=3.2cm]{RESULTS/53215_3.pdf}}
    \caption{ Errors Distribution (CED) curves for sparse and dense face alignment on AFLW2000-3D. Note that for dense face alignment, PRN  \cite{feng2018joint} can only regress around 45K points, so we only select around 45K points for evaluation, even though our model can output all the 53215 vertices provided by the ground truth. Our model performs consistently better on both 2D and 3D problems when compared to other methods.}
    \label{9}
  \end{minipage}
\end{figure}

\begin{figure}[H] %pic result
  \begin{minipage}{\linewidth}
	\centering
	\subcaptionbox{68 points with 3D coordinates.}
	{\includegraphics[height=3.2cm]{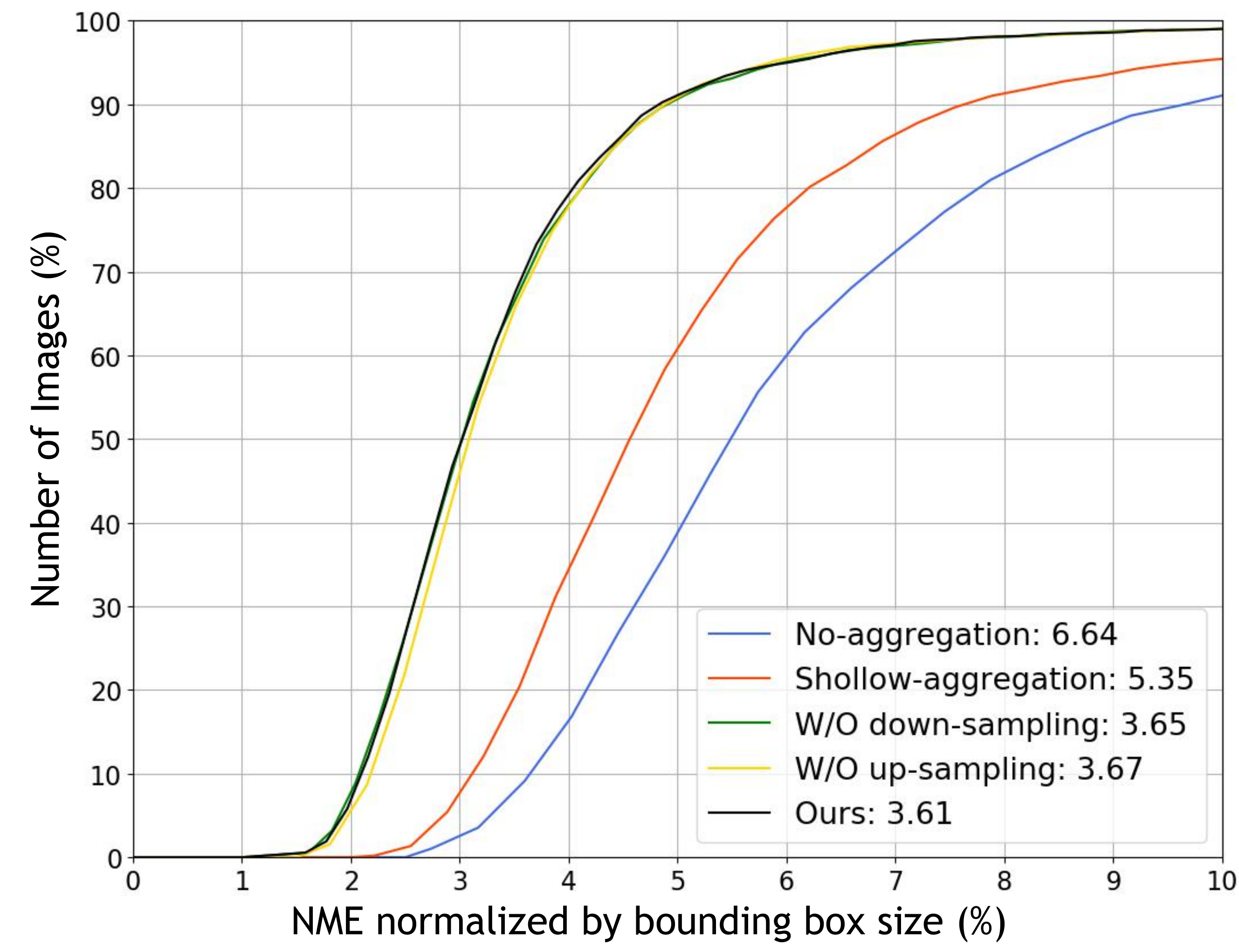}}\quad
    \subcaptionbox{45K points with 3D coordinates.}
    {\includegraphics[height=3.2cm]{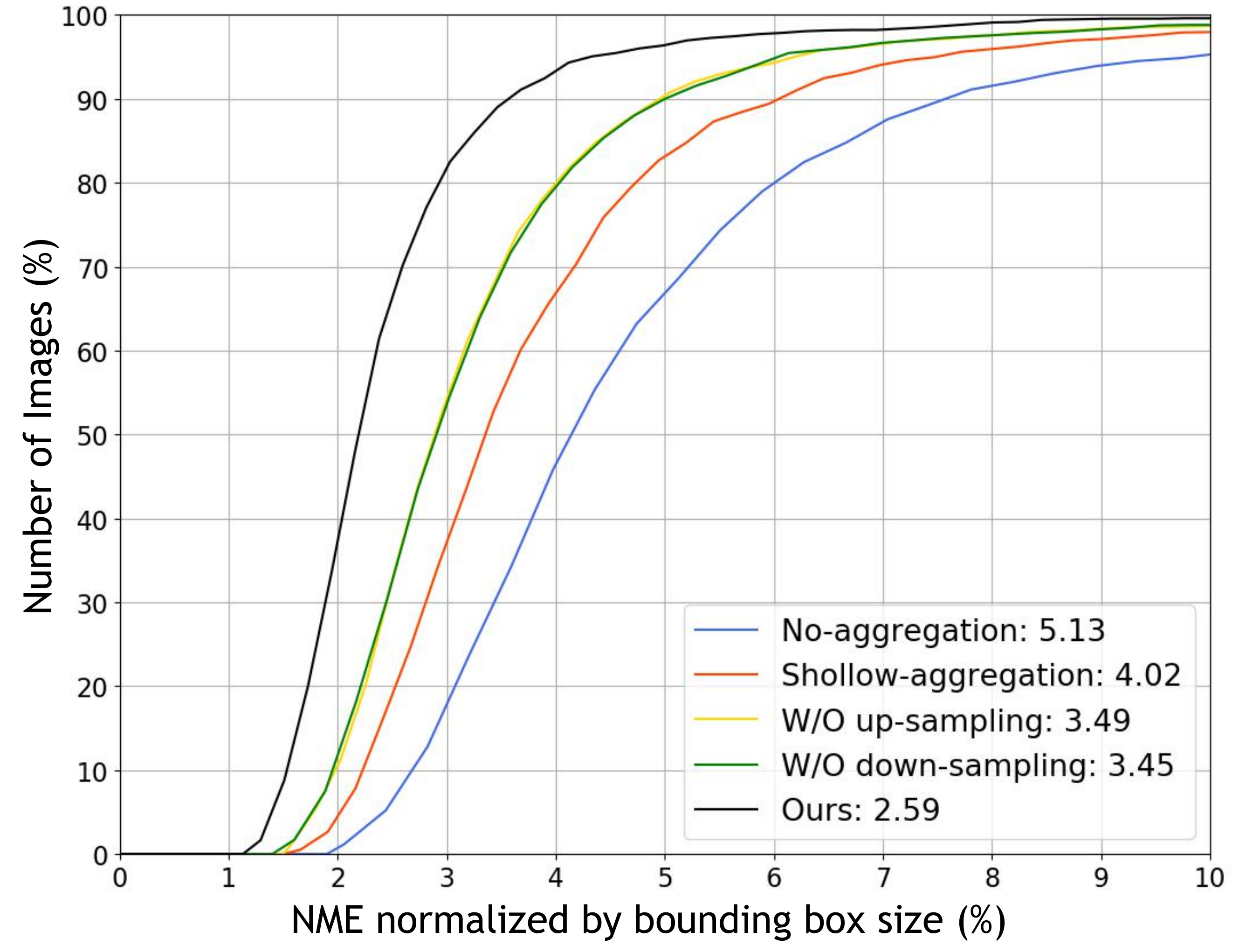}}
    \subcaptionbox{68 points with 3D coordinates.}
    {\includegraphics[height=3.6cm]{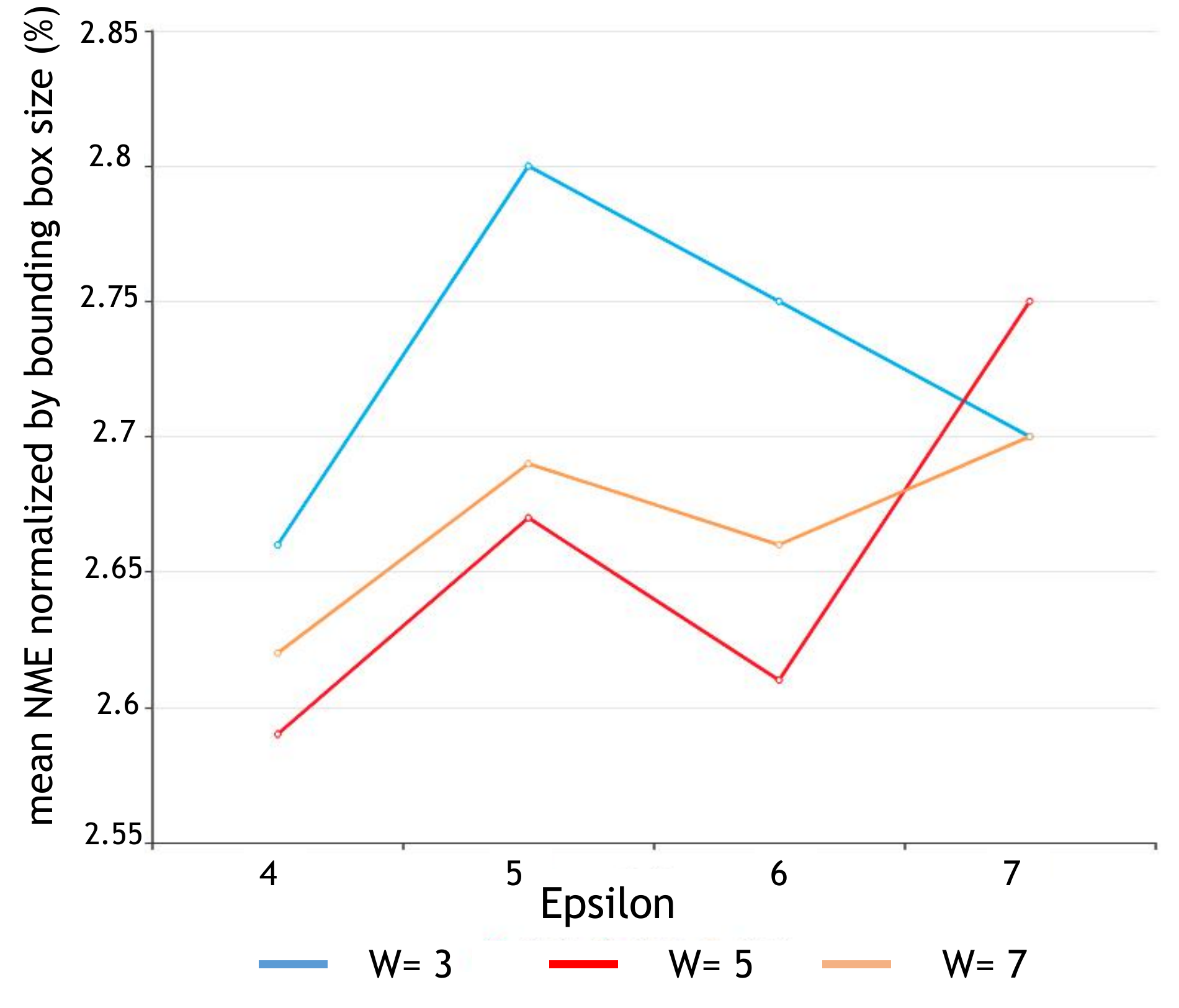}}\quad
    \subcaptionbox{45K points with 3D coordinates.}
    {\includegraphics[height=3.6cm]{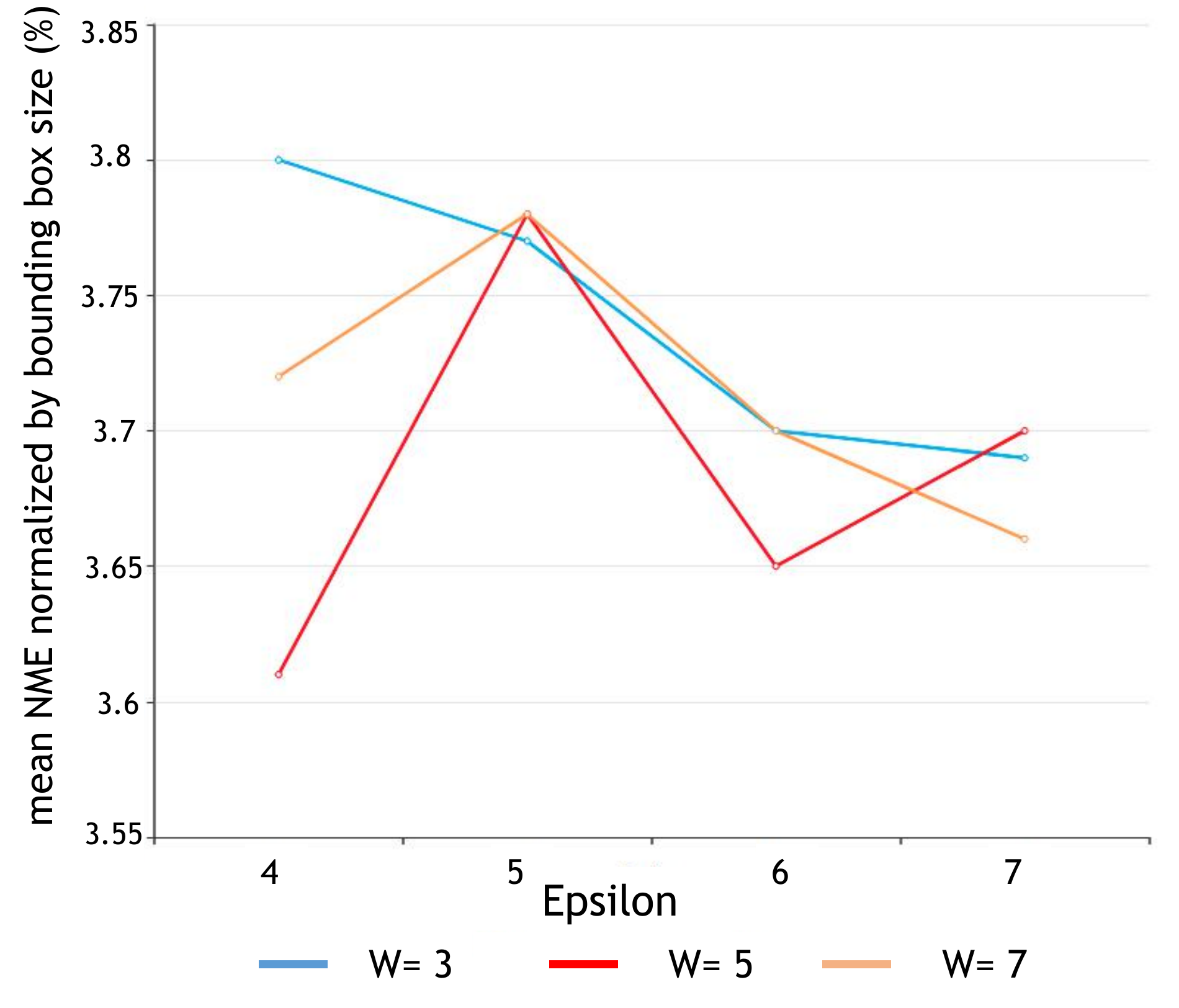}}
	\caption{(a)\&(b), Illustration of the influence of the aggregation block. (c)\&(d), the parameter setting for the proposed loss function. Methods are evaluated on 3D face alignment with 68 landmarks 45K landmarks. Our aggregation model outperforms the other four methods, specifically more than 32\% relative better performance is achieved over the non-aggregation method on both sparse and dense face alignment. And when W = 5, $\epsilon$ = 4, our model achieves best results. }
	\label{12}
  \end{minipage}
\end{figure}

\subsection{Model Complexity and Running Speed}%\vspace{-0.1em}
Even though our model structure looks complicated, in benefit from feature reuse, it is still relatively light-weight and efficient, taking up only 84.5MB compared to 1.5GB in VRN  \cite{jackson2017large} and 153MB in PRN \cite{feng2018joint}. We use the same definition of running time, as suggested by PRN \cite{feng2018joint}. The running time of different models is reported in Tab. \ref{table2}. Our model achieves comparable result with 16.0 milliseconds per image, and the hardware used for the evaluation is NVIDIA GeForce RTX 2080Ti GPU and Intel(R) Xeon(R) W-2104 CPU @ 3.20GHz. The results of 3DDFA  \cite{zhu2016face}, 3DSTN  \cite{bhagavatula2017faster}, CMD  \cite{zhou2019dense} are from their papers, while the running time of the other methods is obtained by running their publicly available source codes on the same machine as our model. %Note that both 3DSTN and 3DDFA use GTX TITAN X GPU to conduct evaluation.
%
%\vspace{-1em}
\begin{table}[h]
	\begin{center}
		\scalebox{0.66}
		{
		\begin{tabular}{|c|c|c|c|c|c|c|c|}
			\hline
			3DDFA  \cite{zhu2016face} & 3D-FAN  \cite{bulat2017far} &PRN\cite{feng2018joint} & DeFA  \cite{liu2017dense} & VRN  \cite{jackson2017large} & CMD  \cite{zhou2019dense} &3DSTN  \cite{bhagavatula2017faster} & \begin{tabular}[c]{@{}c@{}}Ours\end{tabular} \\ \hline
			75.7 ms  & 53.9 ms  & 9.7 ms & 34.5 ms& 68.5 ms& 3.0 ms&19.0 ms& 16.0 ms          \\ \hline
		\end{tabular}
		}
	\end{center}
	\caption{Running time per testing image}
	\label{table2}
\end{table}
%\vspace{-2em}
%\vspace{-2em}
\section{Conclusion}%\vspace{-0.1em}
In this paper, we propose a new end-to-end aggregation graph convolution network to improve the accuracy of dense face alignment and 3D face reconstruction simultaneously. Our network can regress the coordinates of 3D face mesh vertices directly by learning multi-level semantic and spatial features from a single 2D image. Qualitative and quantitative results confirm the effectiveness and efficiency of our model.
%But due to semi-annotated dataset, there may be a potential issue of imperfect ground truth, in the future, our work can be improved by combining with self-supervised approaches.
%It is expected that our model will become a powerful tool for face-related applications and also real-world applications.

% References

\bibliographystyle{IEEEtran}
\bibliography{mybibfile}\ %IEEEabrv instead of IEEEfull

\end{document}